%% file: acl2023.tex
\pdfoutput=1
\documentclass[11pt]{article}

\usepackage{ACL2023}

\usepackage{times}
\usepackage{latexsym}
\usepackage{array}
\usepackage[T1]{fontenc}
\usepackage[utf8]{inputenc}

\usepackage{graphicx}
\usepackage{subcaption}
\usepackage{tabularx}
\usepackage{booktabs}
\usepackage{array}
\usepackage{microtype}
\usepackage{enumitem}

\usepackage{inconsolata}
\usepackage{comment}
\usepackage{hyperref}
\usepackage{amsmath}
\usepackage{cleveref}
\usepackage{xcolor}

\newcommand\todo[1]{{\color{red}TODO:#1}}

\newcommand\bloom{{\texttt{bloomz7b1}}}
\newcommand\mtO{{\texttt{mt0XXL}}}

\newcommand{\xnli}{\texttt{XNLI}}
\newcommand{\xsc}{\texttt{XSC}}
\newcommand{\flo}{\texttt{X$\rightarrow$eng}}

\title{Evaluating Large Language Models along Dimensions of Language Variation: A \textcolor{blue}{Systematik} \textcolor{orange}{Invesdigatiom} \textcolor{magenta}{uv} Cross-lingual Generalization}

\author{Niyati Bafna, Kenton Murray, and David Yarowsky\\
         Johns Hopkins University, Center for Language and Speech Processing  \\
         \{nbafna1,kenton,yarowsky\}@jhu.edu}

\begin{document}
\maketitle
\begin{abstract}
%While large language models are changing the landscape of NLP today, their benefits remain restricted to a few dozens of the world's languages. 
%The study of cross-lingual generalization in large language models (LLMs) is crucial to gaining principle insights for extending LLMs performance to low-resource languages (LRLs). However, we currently lack a fundamental understanding of what kinds of linguistic phenomena are difficult for LLMs in the context of performance on an unseen LRL that is related to a seen high-resource language neighbour (HRLN). In this work, we simulate orthographic/phonological, morphological, and lexical processes to synthesize artificial languages that are controllably distant in these dimensions from seen HRLNs, and study the performance degradation of \bloom{} to varying extents of each process, for machine translation, \xnli, and \xsc. 

While large language models exhibit certain cross-lingual generalization capabilities, they suffer from performance degradation (PD) on unseen closely-related languages (CRLs) and dialects relative to their high-resource language neighbour (HRLN). However, we currently lack a fundamental understanding of what kinds of linguistic distances contribute to PD, and to what extent.  Furthermore, studies of cross-lingual generalization are confounded by unknown quantities of CRL language traces in the training data,
and by the frequent lack of availability of evaluation data in lower-resource related languages and dialects.
To address these issues, we model phonological, morphological, and lexical distance as Bayesian noise processes to synthesize artificial languages that are controllably distant from the HRLN. We analyse PD as a function of underlying noise parameters, offering insights on model robustness to isolated and composed linguistic phenomena, and the impact of task and HRL characteristics on PD. We calculate parameter posteriors on real CRL-HRLN pair data and show that they follow computed trends of artificial languages, demonstrating the viability of our noisers. Our framework offers a cheap solution for estimating task performance on an unseen CRL given HRLN performance using its posteriors, as well as for diagnosing observed PD on a CRL in terms of its linguistic distances from its HRLN, and opens doors to principled methods of mitigating performance degradation.\footnote{\url{https://github.com/niyatibafna/llm-eval-crosslingual-generalization}}
%\footnote{We make our code public here: [anonymized].}
%

\end{abstract}

\section{Introduction}

Advances in the capabilities of large language models (LLMs) have resulted in a paradigm shift in natural language processing, with LLMs being used for and evaluated over a variety of classification and generation tasks \citep{xue-etal-2021-mt5,bang-etal-2023-multitask,hendy-etal-2023-how}. However, even multilingual models such as \bloom{}, \texttt{mT0} \citep{muennighoff2023crosslingual} and Aya \citep{ustun2024aya} only extend model capabilities to $~100$ of the world's highest-resourced languages. The vast majority of the world's $~3800$ written languages have drastically less data available \citep{joshi2020state}, although many have a related high-resource neighbour \citep{asai2023buffet}.
This underscores the need for cross-lingual generalization in LLM capabilities from high-resource languages on which they have been trained to related low-resource languages (LRLs), variants, and dialects, i.e. a theoretical language continuum centered at the high-resource language.

% TODO : 
% (Motivate why we need systematic evaluation of LLMs along dimensions of linguistic change)
% - There are many dimensions of cross-lingual distance, such as phonological/orthographic, morphological, and lexical and we don't know which of them LLMs are capable at, or their error modes with increasing amounts of each phenomenon.
% - We would also like to be able to diagnose performance deterioration on unseen languages based on the above
% - Given an unseen language for which we don't have task evaluation datasets, we would like to able to estimate model proficiency in it a priori, based on its linguistic relationships with its seen high-resource neighbours. We would also like to compare two CRLs in their distance to their HRL neighbour, and understand relative LLM performance degradation as a function of aspects of this distance; this is relevant when we have task datasets for some CRLs but not others.
% The above insights contribute to a principled understanding of the cross-lingual robustment of LLMs, pave the way to targeted interventions to combat performance degradation on low-resource languages.
\begin{figure}
    \centering
    \includegraphics[scale=0.4]{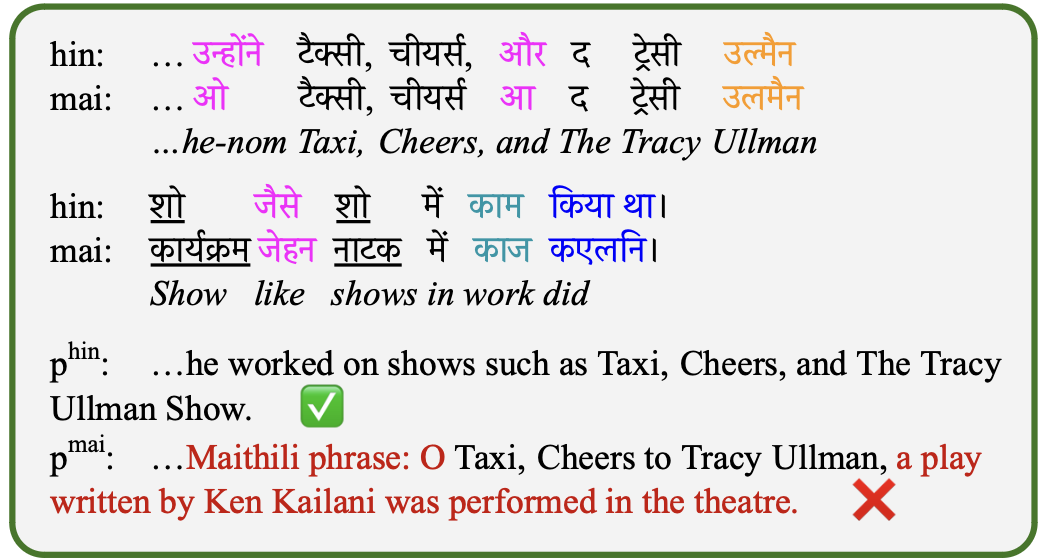}
    \caption{{\color{orange} Phonological/orthographic}, {\color{blue} morphological}, and {\color{magenta} function} and {\color{teal} content} word variation, and \underline{lexical choice} difference, between \texttt{hin} and \texttt{mai}; $p^*$: \bloom{} MT output.}
    \vspace{-12pt}
    % VZ: Use PDF instead of PNG. (2) Please change the neon green. Tame all the colors. (3) Make it wider and shorter. (4) Possibly drop the green stroke but instead just add a light grey rectangle underneath.
    \label{fig:intro_figure}
\end{figure}

Previous literature has reported evidence of multilingual and cross-lingual zero-shot capabilities in LLMs for a number of tasks, also finding, unsurprisingly, that model performance suffers in such settings \citep{jiao-etal-2023-chatgpt,cahyawijaya2024llms} (see \autoref{fig:intro_figure}). While it's reasonable that the farther a closely-related language (CRL) is to its high-resource language neighbour (HRLN), the greater the performance degradation (PD) in a zero-shot setting, we lack a principled understanding of how much different dimensions of linguistic distance (phonological, morphological, and lexical) affect PD. Given that we can find a systematic relationship between each such dimension and PD, and compute the associated distance between a CRL-HRLN pair, this insight would allow us to (a) diagnose observed PD on a CRL, (b) estimate PD for a CRL without task data, as well as (c) suggest targeted interventions aimed at mitigation of PD.

% ----------
%However, we do not currently have a good understanding of the failure modes of LLMs in the context of cross-lingual transfer, i.e. whether LLMs are capable of extrapolating from shared stems, guessing the meanings of new words, or looking past orthographic variations or sound change, and to what extent. Further, we do not know whether LLM performance in evaluated LRLs is a function of cross-lingual generalization or whether the models have happened to see that language or that data in that pretraining corpus; thus, it is unclear whether this performance will hold on truly unseen languages. These insights stand to provide us with a clearer intuition about the best way forwards for extending the advances in LLMs to LRLs depending on its linguistic characteristics and relationships with high-resource pretraining languages, as well as advancing scientific enquiry of the mechanisms of cross-lingual transfer within LLMs.
% -----------

% TODO :
% Why work with artificial languages?

% VZ: manual break because LaTeX doesn't know this word
In this work, we model phonological/ortho-\linebreak
graphic,
%\footnote{While these are inherently different processes, we treat them together since they both manifest as character-level differences on the textual surface form of the language. Orthographic variations reflect minor variations in pronunciation, or phonologically plausible alternations.} 
morphological, and lexical distance as cross-linguistic ``noise'', generated by Bayesian processes applied on a source language, thus positing a parametrization of the HRL dialect continuum.
% for $7$ source languages
% on these languages
We generate artificial languages with varying extents of each noise type, and study LLM zero-shot cross-lingual generalization for three NLU-focused tasks. We discuss the effects of task, noise type, and language family on PD. Crucially, our noise generation processes have tractable posteriors cheaply computable from bilingual lexicons/bitext. This allows us to place real CRLs within the parametrized dialect space of a HRL. We show that PD on real CRLs given their posteriors follows expected trends observed over artificial languages, demonstrating that our noise processes capture useful information about the factors of linguistic distance as they contribute to PD.
% VZ: this just says that you do analysis on your results. That's for granted
% , 

Our use of artificial languages allows us to systematically populate the dialect space of an HRL; further, the noise generation process produces task datasets for each hypothetical language. This solves three problems: firstly, we often do not have task data for real closely-related languages that are unseen in our LLM; secondly, we may not have enough CRLs per HRL, especially CRLs of varying distance along each dimension of interest, to be able to establish and study systematic trends for that language family. Further, we are not guaranteed that a given CRL or its task data is entirely unseen from the training data, confounding a study of LLM zero-shot generalization. Our main contributions are as follows:

% Further, we can now identify cases where language contamination benefits real CRLs, since we have the counterfactual, i.e. truly unseen languages equally distant from the HRL.

\begin{itemize}[left=0mm]
    \item We study the dimensions of linguistic distance that make an input closely-related language difficult relative to its high-resource language neighbour for an LLM in zero-shot settings, quantitatively and qualitatively describing model robustness to each dimension, and discuss the relevance of the task under consideration and the typology and resource-level of the language.
    \item We introduce a parametrization of the dialect space of a language along three linguistic axes that allows for the generation of artificial languages given a set of parameters, as well as for cheaply computing the parameters of a real language pair.
    We demonstrate its utility for predicting and analysing LLM PD on unseen languages using real CRL-HRLN pairs.
    Our framework also opens pathways to
    %to estimating as well as diagnosing PD for a CRL relative to its HRLN, and 
    mitigating PD on low-resource languages, e.g., by reducing damaging distances using linguistic or other tools.
\end{itemize}

%Our work has the following contributions: 1) We show the robustness of LLMs to different dimensions of linguistic distance, as well the effect of the task paradigm being studied 2) We set up a framework for a principled linguistic analysis of why an LRL, or a specific input, may be easy or difficult for an LLM. 3) Our findings propose a clear path forward for work in making difficult LRLs easier for the LLM by transformations in the input space using linguistic tools such as dictionaries and morphanalysers.

%We work with Hindi, Spanish, and Indonesian; these languages all have several low-resource closely related languages  that would benefit from cross-lingual transfer in state-of-the-art language technologies.\footnote{Hindi: the 40+ languages and dialects on the North Indian dialect continuum; Spanish: Galician, Catalan; Indonesian: 100+ creoles and code-switched and dialectical variants spoken in Indonesia.} We work with \bloom{} as a publicly available multilingual model, also presenting baselines results in \texttt{mt0}.

\section{Modelling linguistic variation}
\label{sec:method}
% VZ: manual break because LaTeX doesn't know this word
We model \textbf{p}honological/orthographic, \textbf{m}orpho-\linebreak
logical, and lexical (\textbf{c}ontent and \textbf{f}unction word) variation as parametrized probabilistic ``noisers'' applied to a source language to generate related languages. We denote a noiser as $\phi^n_{v}$, parametrized $\theta_n = v$, where $n \in \{p, m, c, f\}$  indicates the noise type.
For every language, task, and $\phi^n$, we are interested in the function $\psi^n_*: \theta_n \rightarrow PD$,  where
\begin{equation}
  PD = \frac{(s_{\theta}-s_\mathrm{rand})-(b-s_\mathrm{rand})}{b-s_\mathrm{rand}}  
\label{eq:pd}
\end{equation}
Here, $s_{\theta}$ is the performance on the noised source, $b$ is the score on the clean source, and $s_\mathrm{rand}$ is the random baseline.\footnote{$0$ for \flo{}, $33.33$ for \xnli{}, $50$ for \xsc{}; i.e. if \xnli{} score drops to $33.33\%$, we say that it shows $100\%$ PD.} This notation extends to composite noisers, e.g. $\psi^{m,c}_{0.5,*}$ computes PD as a function of $\theta_c$, given $\theta_m = 0.5$. See examples of the outputs of our noisers in \autoref{tab:error_types} and \autoref{app:noising_examples}. 
%% FN We extend this..

% VZ: I don't think this is relevant
% Our code allows for the easy integration of any new noisers.

\subsection{Noiser details}
\label{sec:noisers}
\paragraph{$\phi^p$: Phonological/Orthographic} This model mimics sound change in closely related languages, and is based on the following ideas from theories of sound change \citep{joseph2003handbook}: (i) Sound change is applied to a phoneme given some phonological left and right context e.g. \texttt{(d |a\_,\_EOW)}$\rightarrow$\texttt{t)}. (ii) Sound change, given context, is regular: it applies consistently in all words of the language.
(iii) Consonant sound change largely occurs between phonologically similar phonemes (e.g. difference in voicing: \texttt{f$\rightarrow$v}).
This is not relevant for vowels, which change fluidly. 

% We operationalize these ideas in the following way.
%Note that while our procedure is inspired by these ideas, our goal is to simulate noise that roughly resembles the effects of linguistic sound change and orthographic variation for the purpose of stress testing models nope too defensive
We use manually constructed character$\rightarrow$IPA maps to obtain a set of potential underlying phonemes for script characters.
%\footnote{This is straightforward for languages with phonetically transparent orthographies such as Hindi; we list all possibilities when a character has multiple target phoneme possibilities.} 
%We invert this map to go in the opposite direction.
For any given occurrence of a character, we make a random guess for its corresponding phoneme if there are several.\footnote{Since our goal is to inject random noise into the input roughly guided by the underlying phonology of the text, we can tolerate the imprecision introduced by this process.} We model phonological context as the left and right character of the source character (including word boundaries); thus, a \texttt{(phoneme, context)} pair is simply a character $3$-gram. Each \texttt{(phoneme, context)} is affected with probability $\theta_{p}$.
In order to find a phonologically plausible target set for each IPA character, we construct a list of IPA character sets covering all phonemes used by the languages in this study, such that the phonemes in each set differ from each other in roughly one (or at most two) phonological features, and a phoneme can plausibly change via sound shift to another phoneme in any of the sets it belongs to. (See \autoref{app:phon_noiser}.)
Our list is inspired by \href{https://chridd.nfshost.com/diachronica/full-table}{Index Diachronica}.
% ; note that a single phoneme may belong to multiple sets.}
We can now find a plausible replacement for a given character by mapping it into IPA, sampling a replacement IPA character, and mapping the IPA back into the relevant script. The change to a character given context applies globally throughout the text.

%* Local processes can be modelled using another Bernoulli parameter $p_{pl}$. We model insertions, deletions, and swaps, where the inserted or swapped character comes uniformly at random from the equivalence class of the affected character.
% The relationship between $p_{pg}$ and $p_{pl}$ is not very clear, although in general we imagine a positive correlation for real-world language pairs, and that $p_{pg} < p_{pl}$.

\paragraph{$\phi^m$: Morphological}

Our noiser models concatenative suffixation guided by the following intuitive premises. (i) Affixal change is global (iii) The replacement suffix must be plausible for the language family in terms of its phonology and script, and the original suffix, e.g. if one of them starts with a vowel, the other one is also likely to have an initial vowel. 
We approximate a set of linguistic affixes by collecting the $k$\footnote{empirically chosen per language, e.g. $k=150$ for \texttt{hi}.} most common string suffixes of content words in the language corpus.
%We also apply the heuristic that a suffix is usually not longer than half of a word.} 
Each collected suffix is noised with probability $\theta_m$, by passing it through the phonological noiser as described above, with a high dial ($\theta_p = 0.5$); this ensures the plausibility of the noised target suffix. Finally, we construct a vocabulary map by swapping out all occurrences of an affected source suffix with its generated target in all source words; the vocabulary map applies globally for every occurrence of the word in the text. 

% Note that we do not directly model differences such as changes in case systems, number of genders, inflectional/derivational paradigm differences, but assume that all of these underlying processes manifest on the surface level as affix variations, which can therefore be considered a proxy for morphological variation.

    %\footnote{If a word displays more than one suffix as belonging to our list - this may happen due to noise in our collected list - we choose its ``true'' suffix by sampling from possible suffixes weighted by their global frequencies.} 
    %we ensure that the suffix is noised by at least one character. 
    % \item We now want to sample the suffixes that will be replaced. We would like to bias the selection in favour of commoner suffixes: plausibly, these change first. We use a log frequency weight for every suffix.
    % We proceed in the following way: we first sample the number of suffixes that will be swapped from a $n_{swaps} = binomial(total, p\_{morph}$ distribution. Note that here, $total$ is the (rounded) sum of weights. We then select $n_{swaps}$ randomly from the list of suffixes with their corresponding weights. \textbf{Turns out this posterior has a non-trivial posterior computation, so we'll maybe skip the weights}
    % \item \textbf{For every suffix, we generate a replacement suffix from our character language model. We condition it on the most common observed prefix to encourage the replacement suffix to be a plausible continuation.} (Currently, we're taking maxes instead of sampling from the CLM distributions for each character).

\paragraph{$\phi^{f,c}$: Lexical}

% Functional words: phonological change, high constant dial.
% Content words: Poisson with mean length of original. Generated by char LM.

We model function word change and non-cognate content word change separately, guided by the following premises: (i) The replacement non-cognate equivalent for a content word must be plausible in the relevant script, may not resemble the original word at all, and must not be a word in the source vocabulary. Note that we only model complete lexical change and not lexical choice differences: i.e., when languages have different usage patterns or show semantic shift for the same words. (ii) The length of the replacement word may loosely depend on the length of the original word (for example, words with rare semantics may be longer in both dialects). (ii) Function words in related languages are probably distant cognates, very similar in length.

We identify function words in the input using a list of words appearing with relevant UPOS tags in the Universal Dependencies corpus \citep{nivre2016universal} for each language. Note that since functional words are relatively few and highly frequent, collecting them even over small corpora will yield almost perfect coverage for a given language. Any word not in this list is treated as a content word.

For content words, we sample the length of the replacement word from a $\mathrm{Poisson}(\lambda{=}l)$ where $l$ is the length of the source word, and use a character 3-gram model trained on the language task corpus to generate plausible non-words of the required length. For function words, we generate a replacement by applying a high degree of phonological noise to the functional word ($\theta_p= 0.5$). All replacements for content and function words are global.

% tags: ['ADP', 'AUX', 'CCONJ', 'DET', 'PART', 'PRON', 'SCONJ']

% We operationalize these ideas in the following way:
% \begin{itemize}
%     \item For content words, we sample the length of the replacement word from a $Poisson(\lambda = l)$ where $l$ is the length of the source word. We use a character ngram model with $n=3$ trained on the language corpus to generate plausible non-words of the required length. 
%     \item For function words, we generate a replacement by applying a high degree of phonological noise to the functional word ($\theta_p= 0.5$). All replacements for content and function words are global.
% \end{itemize}
We study lexical change as a combination of $\phi^c$ and $\phi^f$. Since content word change is the more dynamic of the two, likely to show variation depending on language distance, whereas function word change is likely to be high even for related dialects, and show less variation for differently distant languages, we primarily study the PD dynamics of $\phi^{f,c}_{\theta_f,*}$. We experiment with varying $\theta_{c}$, given $\theta_{f} \in \{$0$, $0.5$, $0.8$\}$ ($\phi^{f,c}_{\theta_f,*}$), and with varying $\theta_{f}$ given $\theta_{c}=0$ ($\phi^{f,c}_{*,0}$).

\paragraph{Composite} We compose noisers by independently applying phonological, morphological, and lexical noise in this order (allowing ``overwrites''). While this is a simplification, it is well-motivated; lexical noise is often the most dynamic and continuous of the three while phonological and affixal change are much more gradual and/or fixed given a time period.

\begin{comment}

\paragraph{Syntactic variation}

We model syntactic variation as a Bernoulli over four syntactic properties: \texttt{\{SOV, AdjN, DetN, AdpN\}}. The changed word order is sampled uniformly at random from non-identical possibilities. (This is only relevant for \texttt{SOV} where there are more than one non-identical variations.) We obtain a parse for the source sentence, and then reorder the sentence according to its new target syntax. This can be done both locally and globally.

\subsection{Joint modelling}

* Real-world language pairs can obviously display several of these differences at once. We consider the following combinations...
\end{comment}

\subsection{Posterior computation}
\label{sec:posterior_comp}
We now demonstrate the utility of our noisers and associated $\psi^n$ in understanding PD on real linguistic variation. We assume that  CRLs are ``generated'' by applying a composition of noisers on the source language. Now, if we can find the underlying $\theta_n$, we can estimate $PD = \psi^n_*(\theta_n = v)$, and therefore task performance.

%Given that we trends of performance degradation over the noiser parameter space for a given source language, noise type, and task, we can now use these trends to estimate the performance of our model for an unseen language by computing the parameters that generated it. 
%Note that real language pairs exhibit these noise type simultaneously; we would like to attribute a given change only to only noise type so that we can compose noise types in a meaningful way. 

Given a bilingual lexicon in the source and target, we use word alignments to estimate the Bernoulli parameter $\theta$ $\in$ \{$\theta_p$, $\theta_m$, $\theta_{c}$, $\theta_{f}$\}. In our noisers, all changes to the concerned units (trigrams, suffixes, words) are global. In reality we may not observe a global change between source and target unit; language change may be noisy, we may have one-off phenomena, and we may have noisy word alignments. We compute $\theta$ in the following way:
\begin{gather}
E[\theta] = \frac{\sum_u I_u}{T} , \qquad E\left[\frac{\sum_u I_u}{T}\right] = \sum_u \frac{E[I_u]}{T}
\nonumber
\end{gather}
where $I_u$ is a binary random variable indicating whether unit $u$ was affected, and $T$ is the total number of units. We can now estimate $E[I_u] = \frac{C_u}{T_u}$ for each $u$ i.e. the fraction of times that $u$ was affected. Note that it remains to be decided how we will categorize a given change in a non-identical source-target pair. 

\paragraph{Phonological} If source-target normalized edit distance (NED) is high,\footnote{We use language-specific empirically determined thresholds for NED-based decisions, e.g. $0.5$ for \texttt{de} in this case} we attribute changes in the target word to phonological change. We find the minimal list of edits from source to target; if we observe a character change with the same left-right context, we count it towards $\theta_p$.

\paragraph{Morphological} If a content target word has a different suffix (identified as in \autoref{sec:noisers}) but the same stem but (i.e. it is not lexical change)
%\footnote{Suffixes are identified using our collected list as in \autoref{sec:noisers}.} 
, we count it towards $\theta_m$.
% REMOVE identified as in...

\paragraph{Lexical} We count any change in a function word towards $\theta_{f}$ . For content words, if the source-target NED is low (i.e. not phonological/morphological change) and the target word is not present in the source vocabulary, we count it towards $\theta_{c}$.

While real languages exhibit the above kinds of noise simultaneously, i.e. they are the result of composite noising, our model of noise composition permits us to compute the posteriors of individual noisers independently of each other. Note that allowing overwrites by successive noisers does not affect this property: although lexical change may ``overwrite'' a suffix change, it does not change the fraction of suffixes/trigrams affected (i.e. the MLE estimate of $\theta_m$), since the noisers are independent of each other.\footnote{We compute $\theta_{m}$ only over words that have the same stem in source and target; any word pair with different stems is ignored. Since lexical noise is applied uniformly over words and independently of morphological noise, we expect that while it will ``disqualify'' a set of word pairs for the $\theta_m$ posterior computation, the remaining set will give us the same estimate (in expectation) of $\theta_{m}$. An analogous argument applies for $\theta_p$.}

\section{Experimental Setup}
\label{sec:exp_setup}
\paragraph{Model and Tasks}
% \subsection{Models}
%* mGPT, GPT-3.5, BLOOMZ
% We computed baseline results for \texttt{mGPT}, \mtO, and \bloom over our selected tasks, reported in Appendix \todo{put baseline results for all models}. We note that \mtO and \bloom consistently outperform \texttt{mGPT} but show comparable performance with each other. We primarily with \bloom for our experiments, and additionally replicate translation experiments on \gpt.\footnote{We could not evaluate \gpt on the other tasks since it does not allow access to loglikehoods of input tokens.}

We obtain initial zero-shot results on a number of tasks for \bloom and \mtO{} \citep{muennighoff2023crosslingual}, and select three tasks to work with: \flo{} machine translation on \texttt{FloRes200} \citep{nllb2022},\footnote{We loosely refer to \flo{} as an NLU task; since the LLM is fluent in English, its performance primarily depends on comprehension of the input \citep{nguyen-etal-2024-democratizing}.} \texttt{XStoryCloze} \citep[\xsc;][]{xstorycloze}, and \texttt{XNLI} \citep{conneau2018xnli}, as covering a large enough mutual set of languages as well as two tasks paradigms of interest, namely, multiple-choice questions and sequence-to-sequence. We found that the performance of both models on multilingual ARC, HellaSwag and MMLU \citep{dac2023okapi} is close to or worse than chance for many languages; this makes these tasks unsuitable for studying model PD. 

Our experiments are conducted on \bloom{}, using the \texttt{mlmm-eval} evaluation framework \citep{dac2023okapi}. See §~\ref{app:baselines} and §~\ref{app:prompts} for all evaluated tasks and further experimental details.
% for details of prompt choice and other experimental details.

\paragraph{Languages}
We work with Hindi, Indonesian, Arabic, German, French, Spanish, and English. This set of languages was curated with language presence in \bloom{}\footnote{German is ``low-resource'' for \bloom{}, constituting only $0.21\%$ of the training corpus \citep{muennighoff2023crosslingual}.} and availability of task datasets in mind. We include three macrolanguages (\texttt{hi}, \texttt{id}, 
\texttt{ar}) with dozens of real closely related low-resource languages and dialects. In order to validate our computed trends with real language data, we require languages and dialects related in varying extents to the respective HRLN, unseen from \bloom{}, with task dataset availability. We study trends for \flo{} for the following CRLs: Awadhi-\texttt{awa}, Bhojpuri-\texttt{bho}, Magahi-\texttt{mag}, Maithili-\texttt{mai}, and Chhattisgarhi-\texttt{hne} (Hindi), Danish-\texttt{dan}, Icelandic-\texttt{isl}, and Swedish-\texttt{swe} (German), Malay-\texttt{zsm}-(Indonesian), Occitan-\texttt{oci} (French), Galician-\texttt{glg} (Spanish), and Iraqi-\texttt{acm}, Yemeni-\texttt{acq}, Tunisian-\texttt{aeb}, Levantine-\texttt{ajp}, North Levantine-\texttt{apc}, Najdi-\texttt{ars}, Moroccan-\texttt{ary}, and Egyptian-\texttt{arz} (Arabic). This list includes language pairs with a range of degrees of relatedness; e.g. \texttt{zsm} and \texttt{ind} are much closer than \texttt{dan} and \texttt{deu} \citep{dryer2013world}. 
%we work with Awadhi, Bhojpuri, Magahi, Maithili, and Chhattisgarhi (Hindi), Danish, Icelandic, and Swedish (German), Malay (Indonesian), Occitan (French), and Galician (Spanish), for \flo{}. We obtain bilingual lexicons from Google Translate when available, and alternatively use statistical word alignment with FastAlign \citep{dyer2013simple} on FloRes bitext.\footnote{We manually filter $300$ entries for \texttt{mai} and \texttt{hne} and verify that the computed posteriors over possibly noisy alignments are similar to those computed on clean lexicons (see \autoref{app:posterior_details}).}

\input{error_types}

\section{Results and Discussion}
\label{sec:discussion}
\begin{figure*}
    \centering
    \includegraphics[scale=0.38]{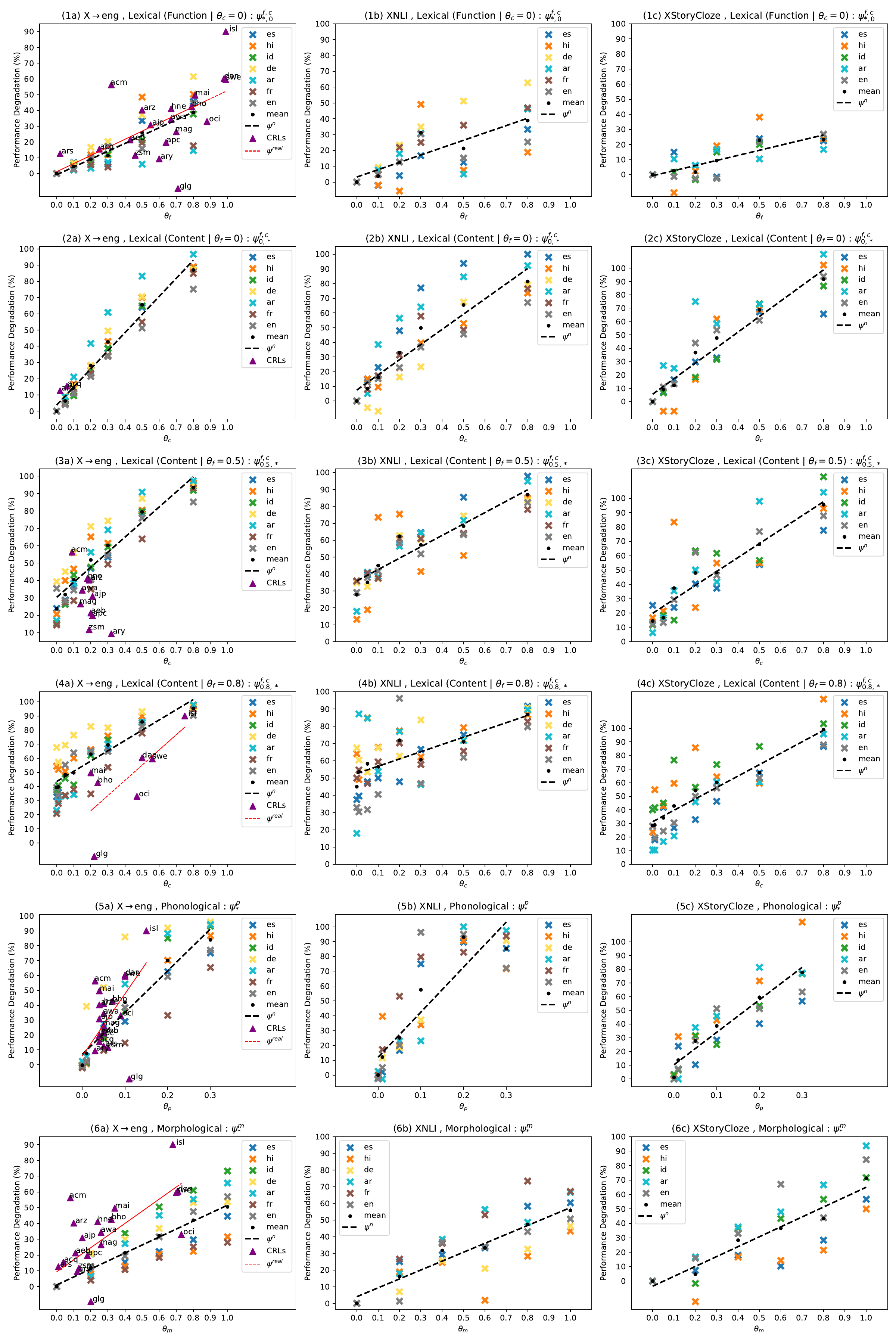}
    \vspace{-3mm}
    \caption{PD\% for each language, task, and noiser. $\psi^n$ depicts mean language PD trends. We show $(\theta_n, PD\%)$ points for real CRL-HRLN pairs using computed posteriors for \flo{}. See \autoref{sec:exp_setup} for corresponding HRLNs per CRL. $\psi^{real}$ depicts trends for real CRLs, shown only when $\theta_{real}$ has a wide-enough range.}
    \label{fig:results}
\end{figure*}

See $\psi^n$ for noiser, task, and language combinations in Figure~\ref{fig:results} (single run per noiser parametrization). 
%Also see input-output examples in Table~\autoref{tab:outputs_examples}.
\paragraph{Tasks} 

We find that \textbf{the rate of mean PD given a noise type is the same across tasks}. \textbf{This indicates that model performance for one task for a CRL relative to its HRLN can be used to extrapolate its performance on other tasks; i.e. PD is largely a function of language distance.}

While we see linear trends for mean PD for all tasks and noise types, and individual languages trends are also linear for \flo{}, this is less true for individual language trends for \xsc{} and \xnli{} (e.g. 3b, 3c, 4b, for \texttt{arb,hin}). This is a result of sampling variance in our noising process: $\phi^n_v$ may produce a range of artificial languages varying in the specific set of units that are noised. The relationship between PD and $\theta_n$ is mediated by task sensitivity to the comprehension of specific words (phones/morphs) as opposed to general comprehension of the input: we compute std. deviation of PD for multiple artificial languages generated from the same $\theta_n$ for \texttt{hi} and \texttt{ar}, and find much lower SD for \flo{} than the other tasks. Using PD means over multiple artificial languages per $\theta_n$ removes the instability of the trend at the individual language level and is key to computing reliable trends for a language. See \autoref{app:trend_stability} for task-wise std. dev. and stabilized trends for \texttt{hi} and \texttt{ar}.

These findings back the intuition that while translation depends on local understanding of input, suffering predictably with increasing noise, the model relies only on certain words rather than the entire sentence for classification tasks, and is therefore more sensitive to whether those  are corrupted rather than the general extent of noise, although of course these two are correlated. \xnli{} in particular is also highly sensitive to whether its three label words are noised, strongly cautioning any zero-shot evaluation to be mindful of its treatment of label words. \textbf{This suggests that \flo{} is a more robust test of NLU in a LRL for a model, and less susceptible to fluke performances.}

% for \texttt{hi, ar}, $\phi^{f,c}_{v,*}$, \xnli{}.
 % , and underlines the fragility of such tasks.

%- Optional: (maybe implied by above and below?) Higher spread across languages for XNLI and XSC for other tasks than translation. (verify)
% - This is actually not true. Variance of PD across languages seems to be a function of noise type.
%Compute variance across languages given task over all noise types

\paragraph{Languages}

\begin{figure}[!ht]
    \centering
    \includegraphics[scale=0.7]{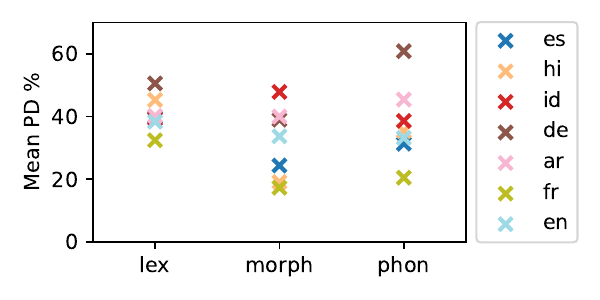}
    \caption{Mean PD over all parametrizations  per noiser for \flo{}}
    \label{fig:mean_deg}
    \vspace{-8pt}
    % VZ: This figure can be half the height.
\end{figure}

We see that \texttt{ar} and \texttt{id} suffer most from $\phi^m$ (e.g. 6a), perhaps due to their rich morphology \citep{lopo2024constructing}, and that \texttt{de} particularly suffers from $\phi^{c}$ (e.g. 4a), possibly because word compounding results in a higher extent of lost information per noised word. See \autoref{fig:mean_deg} for mean PD over all parametrizations of a given noiser per language for \flo{}. In general, we find that lower-resource languages in \bloom{} such as \texttt{de}, \texttt{ar}, \texttt{id}, and \texttt{hi} have higher mean PD as compared to HRLs like \texttt{fr} and \texttt{es}; \textbf{more exposure to a language makes the model more adept at unseen related languages.}

% Get figure from: mean_degradations.png in analysis/plots/

\paragraph{Noise types} The slope of $\psi^n$ signals how damaging noise type $n$ is (higher is worse). We contextualize these trends over $\theta$ using the posteriors computed over real language pairs, which provide a sense of the natural range of $\theta$ for related languages per noiser. Note that absolute PD values for a given $\theta_n$, and therefore absolute slopes, are not comparable across noise types, since $\theta_n$ differs in meaning depending on the noiser; however, these can be compared directly for different lexical noisers.
%%FN Note that...

We find that $\phi^{f,c}_{*,0}$ shows lower PD rate as compared to $\phi^{f,c}_{0,*}$: naturally, \textbf{content loss is more damaging than function word loss}. However, note that real $\theta_{f}$ values are high even for very closely related language pairs (e.g. \texttt{hne-hin}; see 1a), and correspond to significant PD values. On the other hand, $\theta_{c}$ may be low ($<0.2$) for closely related languages, but is more costly. Note that $\psi^{f,c}_{\theta_f,*}$ for $\theta_f \in \{0,0.5,0.8\}$ have similar slopes but increasing $y$-intercepts based on $\theta_{f}$. Given that function words form a closed and relatively small set for a given language, and may be easier to deal with than open class, possibly rare, content words, \textbf{this suggests that we can cheaply tackle a non-trivial portion of PD by simply handling ``easier'' function word correspondences}. 

We observe that $\psi^m_*$ displays a low slope; corrupting $100\%$ of our set of linguistic suffixes results in a mean $50-70\%$ PD. \textbf{This indicates that the model is largely capable of capturing important information from word stems.} Note that for distant related cousins like \texttt{de-dan}, $\theta_m$ can be high and correspond to significant PD. 

Finally, $\psi^p_*$ indicates sharp PD; this is natural since $\phi^p$ affects chargrams with possibly widespread effect in the corpus. Once again, while our chosen LRLs cover a range of natural values for $\theta_p$, even very closely-related languages display $\theta_p$ values corresponding to significant PD (5a), \textbf{suggesting that the model is vulnerable to natural levels of phonological/orthographic variation}.
%Observed high susceptibility of \texttt{ar} to this noiser can be explained by model unfamiliarity with non-Latin scripts.

%\footnote{e.g. single suffix change may cause sentence-wide hallucination, whereas in other cases the model does well despite many consecutive corruptions (\autoref{app:details}).} 
%However, as $\theta$ increases, we note a natural shift in the output modes towards hallucination and off-target translations.\todo{Figure \% of off target with increasing theta.}

%\paragraph{PD over noise composition} While overall PD for a language with composed noisers is a presumably a function of PD for each contained noise type, the nature of this function remains to be understood. We study $\phi^{f,c,m}_{0.5,*,0.5}$, composing lexical and morphological noise (see \autoref{fig:composite_lex_morph} for \flo{} and \xsc{}) and observe that for \flo{}, the resulting PD is well-explained simply by $\phi^{f,c}_{0.5,*}$; indicating that overall PD may be a simple max (as opposed to incremental) in this case.\footnote{\xnli{} follows this trend; see \autoref{app:pd_composed}.} This idea offers one explanation of the observed PD of \texttt{isl}, i.e. that the PD effect is dominated by $\phi^{f,c}_{0.8,*}$. However, for \xsc{}, we observe that $\psi^{f,c,m}_{0.5,*,0.5}$ in fact exceeds the theoretical additive noise trend. While we leave a detailed study of this composition function to future work, we show that it is task dependent; we also believe that it is likely to be vary depending on noiser combination.

\begin{figure}[ht]
    \centering
    \includegraphics[width=\linewidth]{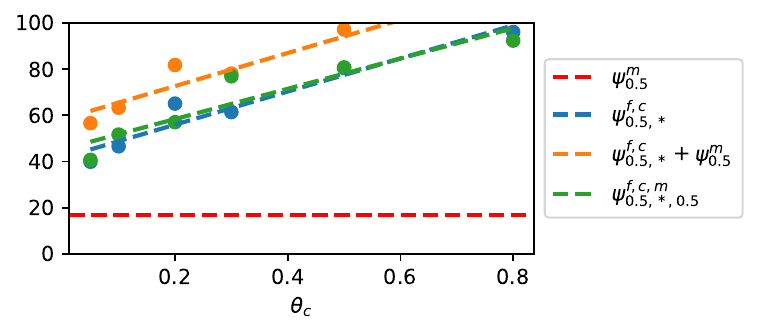}
    \caption{Composing $\phi^{f,c}$ and $\phi^{m}$: studying $\psi^{f,c,m}$ given  $\theta_m$ for Hindi for \flo{}. $\psi^{f,c}+\psi^{m}$ shows the theoretical additive trend.}
    \label{fig:composite_lex_morph}
\end{figure}

\paragraph{PD over noise composition} While overall PD for a language with composite noising is a presumably a function of PD for each contained noise type, the nature of this function remains to be understood. We study $\phi^{f,c,m}_{0.5,*,0.5}$, composing lexical and morphological noise (see \autoref{fig:composite_lex_morph} for \flo{}),and observe that the resulting PD is well-explained simply by $\psi^{f,c}_{0.5,*}$; indicating that overall PD may be a simple max (as opposed to incremental) in this case. 

We can show that within our framework of composition, the effect of a constant amount of morphological/phonological noise decreases as lexical noise grows, due to an increasing “overwrite” probability in composition.\footnote{See \autoref{app:pd_composed} for a formal explanation, as well as for $\psi^{f,c,m}_{0.5,*,0.5}$ for \xnli{} and \xsc{}.} This matches our intuition about linguistic variation well: \textbf{as languages grow lexically distant, lexical change becomes the dominating factor in PD}. This is because growing lexical change transforms more words to non-cognates, rendering the underlying phonological or morphological patterns affecting cognates decreasingly relevant. This idea offers one explanation of the observed PD of \texttt{isl}, i.e. that the PD effect is dominated by $\phi^{f,c}_{0.8,*}$. We leave a detailed study of PD for composite noisers as a function of individual noiser PD to future work. We believe that it is likely to depend on the noiser combination (e.g. $\phi^{p,m}$ vs. $\phi^{p,f,c}$), as well as the comparative initial PD for the isolated noisers: whether we are composing equal or imbalanced levels of noise (or resulting PD) from isolated noisers may influence the nature of composition.

% \begin{figure}[ht]
%     \centering
%     \begin{subfigure}{0.45\textwidth}
%         \centering
%         \includegraphics[width=\linewidth]{images/composite_lex_morph_flores.pdf}
%     %     \caption{\flo{}}
%     % \label{fig:composite_lex_morph_flores}
%     \end{subfigure}
%     \vspace{-10pt}
%     \begin{subfigure}{0.45\textwidth}
%         \centering
%         \includegraphics[width=\linewidth]{images/composite_lex_morph_xsc.pdf}
%         % \caption{\xsc{}}
%         % \label{fig:composite_lex_morph_xsc}
%     \end{subfigure}
%     \caption{Composing $\phi^{f,c}$ and $\phi^{m}$: studying $\psi^{f,c,m}$ given  $\theta_m$ for Hindi for \flo{} (top) and \xsc{} (bottom). $\psi^{f,c}+\psi^{m}$ shows the theoretical additive trend.}
%     \label{fig:composite_lex_morph}
% \end{figure}
% \vspace{-10pt}

\paragraph{Posteriors and trends for real CRLs}

 % on the relevant plot
We calculate posteriors for real CRLs as described in \autoref{sec:posterior_comp}. This procedure requires bilingual lexicons: we obtain these from Google Translate when available, and alternatively use statistical word alignment with FastAlign \citep{dyer2013simple} on FloRes bitext. We verify that computed posteriors over possibly noisy alignments are similar to those computed on clean lexicons by comparing posteriors obtained from noisy and manually cleaned lexicons for \texttt{mai} and \texttt{hne}: we find that this is largely the case for $\theta_f$, $\theta_p$, and $\theta_m$, but that $\theta_c$ is prone to being over-estimated from noisy alignments. 

We plot $(\theta, PD)$ points for \flo{} in \autoref{fig:results}. We bucket the $\theta_{f}$ posterior and show $(\theta_{c}, PD)$ on the relevant $\psi^{f,c}_{\theta_f,*}$ plot. Note that we can use posteriors for a CRL-HRLN pair to generate artificial languages that are equally distant from the HRLN as the LRL; we provide examples in \autoref{app:pseudo-crls} to illustrate the plausibility of our noisers and associated posteriors. 
%\footnote{Note that these datapoints are different from the artificial language datapoints, since real languages contain a composition of all noise types, and performance degradation for these may be a function of all of them.} 
\textbf{We observe that PD vs. $\theta_n$ for real languages generally follow similar trends as $\psi^n$, indicating that our constructed $\phi^n$ offer useful parametrizations of linguistic distance as it contributes to PD.} 

Notable outliers are are \texttt{oci}, \texttt{zsm}, and \texttt{acm} for $\phi^{f,c}$. Further, \texttt{glg} actually performs with $+4$ BLEU over \texttt{es} (\autoref{app:posterior_details}), which is a clear red flag. Such anomalies, where observed PD is much lower than expected PD, could indicate unreported amounts of the language in the training data or, in the case of \texttt{glg}, possibly test set leakage. 

Note that since real languages contain a composition of all noise types, we expect total PD to be higher than that predicted by any individual $\psi^n$. However, this is not true, notably observed for $\psi^{c}_*$ and $\psi^{f}_*$ (3a, 4a).
%For example, we see that \texttt{isl} PD is predicted well, and possibly entirely explained by, $\psi^{c}$ and $\psi^p$, and that $\phi^{m}$ and $\phi^{lf}$, which do not account for extra lexical/phonological noise, widely underestimate it. 
%While $\psi^{m}_*$ and $\psi^{p}_*$ generally underestimate PD in accordance with this intuition, 
This is attributable to code-switching and traces of the unseen language in training data. For artificial languages, the cost of a completely unknown word is high (as compared to a partially known, suffix-corrupted word); however, it's likely that the model actually knows some percentage of words identified as unknown by our posterior computation in the real unseen languages. The unknown word may be present in another language than the HRLN (e.g. \texttt{fr-oci} \texttt{changement-cambiar}; \texttt{cambiar} is a Spanish equivalent), or it may be non-identical but very close to an HRLN synonym (\texttt{certain-qualques} - French synonym \texttt{quelques}), or it may simply be known because the model has seen data in the ``unseen'' language. This would have the effect of reducing the absolute PD while maintaining the trend. \textbf{The observed delta between the trends gives us an idea of the benefits of multilinguality and language contamination in training data by providing the counterfactual.}

See \autoref{app:posterior_details} for more details on the effect of noisy alignments on posteriors, and the computed posteriors, associated BLEU scores, and PD for each $\theta_n$ and CRL.

%, \texttt{regarder-veire} (French synonym \texttt{voir}).
\paragraph{Realistic quality of artificial languages} We provide examples of our generated artificial languages in Tables~\ref{tab:error_types},~\ref{tab:noiser_examples_extra},~\ref{tab:all_error_modes1},~\ref{tab:all_error_modes2},~and~\ref{tab:composite} for various HRLNs. The languages appear phonologically and orthographically plausible given the language family of the HRLN, with occasional transgressions. For example, an artificial language may not respect rules for diacritic vs character usage for vowels in Devanagari, or may over-use rarely observed characters in a script. Currently, we also do not have special treatment for named entities, which should ideally remain unnoised. 

\input{case_study_awa}

\paragraph{Error Modes}
See \autoref{tab:error_types} for a qualitative classification of model error modes for each noiser, obtained via a manual examination of outputs over representative $\theta_n$. See an expanded version of this table in \autoref{app:noising_examples}. We also perform a small case study in error type characterization over 70 sentences for \texttt{awa-eng} \flo{}, shown in \autoref{tab:case_study}. These qualitative analyses indicate that the model is able to withstand a good extent of phonological, morphological, and function word change (exhibited in its $38.5\%$ near-perfect translations for \texttt{awa-eng} MT), but fails in different ways when multiple morphological and function word changes are in close proximity. Entire content word shifts as opposed to lexical choice variation were rare between Awadhi and Hindi in our sample but cause breakage when they occur. We note that these error type and diagnosis distributions will differ based on the language pair under consideration, the nature of the divergence between the CRL and HRLN and their typologies, as well as the LLM proficiency in both. 

In general, while PD over a dataset varies smoothly as a function of $\theta_n$, we observe that \textbf{success/failure modes over individual inputs are not easily predictable: the model displays both surprising fragility as well as robustness in different cases.}

\section{Related Work}

\paragraph{Multilingual evaluation of LLMs} Recent studies show that LLMs demonstrate certain multilingual capabilities accompanied with performance degradation for LRLs for machine translation \citep{jiao-etal-2023-chatgpt,hendy-etal-2023-how,robinson-etal-2023-chatgpt} as well as other tasks like POS, NER, and summarization \citep{lai-etal-2023-chatgpt,bang2023multitask,asai2023buffet}. \citet{kantharuban-etal-2023-quantifying} attempt to identify economic, social, and linguistic correlates of MT performance in LLMs for dialects; they find positive correlations for dataset size and lexical similarity among other factors. It is difficult to draw principled insights from such studies about what the bottlenecks for cross-lingual transfer are, since the tested languages may simultaneously vary in their relatedness to high-resource languages, and presence in the pretraining data. 

\paragraph{Linguistic distance as a factor in performance} Recent work explores providing ``missing'' linguistic knowledge of LRLs (lexical, morphosyntactic) in LLMs by providing dictionaries, bitext, and grammar books via in-context learning for LRLs \citep{tanzer-etal-2024-benchmark,zhang-etal-2024-hire,zhang-etal-2024-teaching}. Other works look at cleverly choosing shots for the context by exploring the prompt space, choosing exemplars that are ``close'' to the output using lexical distance \citep{zhu-etal-2023-multilingual,zhang-etal-2024-teaching,cahyawijaya2024llms}. However, this search space of what can be provided is large, and we lack an understanding of which linguistic distances LLMs need ``help'' with: these ideas motivate a study such as ours.

\paragraph{Robustness} Earlier studies have looked at robustness of machine translation systems to orthographic variants, typos, and other kinds of noise \citep{belinkov-bisk-2018-synthetic,heigold-etal-2018-how}. \citet{moradi-samwald-2021-evaluating} perform a similar study of BERT-like models for sentiment analysis, QA, and NER, among other tasks, with the intent of stress-testing LMs against natural user-generated noise such as synonym replacement, common misspellings, and verb tense errors. \citet{wang2023robustness} discuss the robustness of ChatGPT against adversarial and out-of-distribution input datasets such as ANLI and DDXPlus. 
%\citet{dong-etal-2023-revisit} look at the effect of ASR-created noise for slot filling in a goal-oriented dialog system for English, studying typos, simplifications, verbosity, paraphrasing, and word and sentence-level perturbations. 
\citet{havrilla-iyer-2024-understanding} investigate character-level static and dynamic noise for chain-of-throught prompting processes. As far as we know, ours is the first work to stress test LLMs under noise models of linguistic distance.
%This is the gap that our work addresses by systematically answering questions about the extent of tolerance of LLMs to various kinds of linguistic distances between a seen HRL and an unseen LRL.

\section{Conclusion}
\label{sec:conclusion}
%We study robustness  to linguistic distance in a zero-shot context for $7$ languages and $3$ tasks. 
%We model three aspects of linguistic distance as Bayesian noise processes applied to HRLs and study LLM performance degradation for $3$ tasks and $7$ languages on artificial related languages as a function of noise parameters: this allows us to quantitatively and qualitatively characterize the impact of each isolated linguistic phenomenon. We show that PD for real unseen languages follow expected trends given their computed posterior noise parameters, validating our noiser construction. Our work offers a framework for the principled linguistic analysis of cross-lingual generalization and opens avenues in mitigating LLM performance degradation in low-resource settings.
%\nb{fix}
We study the robustness of an LLM to $4$ types of linguistically-motivated (phonological, morphological and lexical) Bayesian noise models on $7$ languages and $3$ tasks, generating artificially languages controllably distant from a given HRL and computing trends in performance degradation. This allows us to quantitatively and qualitatively characterize the impact of each included factor of linguistic variation on task performance in isolation. Our noisers are amenable to cheap posterior computation; we show that PD for real unseen languages follow expected trends given their computed posteriors, validating our noiser construction. Our work offers a framework for the principled linguistic analysis of cross-lingual generalization and opens avenues in mitigating LLM performance degradation in low-resource settings.

%\newpage

\section*{Limitations}
\label{sec:limitations}

\paragraph{Noiser choice} Our work is limited by the three linguistic phenomena we study. Notably, we do not study syntactic change, since it is not naturally modeled by our framework of smoothly increasing distances in a hypothetical continuum.  This is for mainly two reasons: firstly, there are simply far fewer possible syntactic changes in total (core syntax can be described within 10-15 features); secondly, systematic syntactic change is much rarer in related languages (very few of those features actually change within language families).

It is certainly possible to extend this study to other noisers modeling relevant phenomena in the context of language continua. One example is the phenomenon of semantic shift, whereby words with the same form shift in meaning in related languages, resulting in different lexical choice for the languages (although not lexical change); lexical usage patterns in general may also be of interest. We give an example of this in \autoref{fig:intro_figure}. This can be modeled within our framework as a noiser that moves a word to its synonym with some probability; we leave such ideas to future work.

\paragraph{Noiser design}
Our noisers incorporate several simplifications from a linguistic standpoint. Each noiser can be further nuanced to increase the plausibility of the resulting synthesized languages; some examples of possible detailing include (a) $\phi^p$: using language-family-specific sound change models that weight commonly observed sound changes in that family higher than others (b) $\phi^m$: using morphological tools to more accurately identify linguistic suffixes, (c) $\phi^m$: modeling other kinds of morphology, e.g. non-concatenative, templatic, prefixal. This is particularly relevant to languages such as Arabic. (d) $\phi^c$: introducing weighting by (log) frequency such that commoner words are more likely to be affected by the noiser. Note that some of these changes may introduce complications for posterior computation. We leave it to future work that is interested in particular noisers for particular language families to look into fine-graining noiser design in a given context. 

% Different models, few shot, languages
\paragraph{Comprehensiveness: Languages, Tasks, and Models}
Our insights on PD characterization are limited to the 3 tasks and 7 languages we study, in a zero-context context for \bloom{}. Each of these dimensions can naturally be expanded: it is possible that the observed PD dynamics are different for different models (individual trends for a noiser will certainly differ depending on model, language, and task), or for a few-shot context. We focus on three NLU-oriented tasks for our study; our conclusions about cross-lingual transfer may change for different task paradigms \citep{ahuja-etal-2022-multi}. Further, we are also able to provide our results on real language posteriors only on \flo{}; we are constrained by task dataset availability for truly low-resource languages. We make our code available and encourage a similar analysis to ours for any new combination of language, model, task, noiser, and experimental setting.

% Composition dynamics
\paragraph{Noiser composition dynamics}
Our work focuses mainly on PD dynamics for individual noise types to isolate the effect of each linguistic phenomenon, and touches only briefly on the PD dynamics for composed noisers, although our noise processes and posteriors offer natural extensions for noise composition. While we demonstrate the complexity of observed PD dynamics on a single language and single noise composition setup for 3 tasks, we leave a detailed investigation of the same, which should include a large enough selection of noiser combinations for different language typologies, tasks, and parametrizations per noiser, to future work.

% Error modes on real languages
%Finally, related to the above: while we characterize error modes and provides examples for model outputs on noised inputs for individual noise types, these may be different for composed noisers, and by consequence, for real languages.

\section*{Ethics Statement}

Our work is motivated by the need to increase language inclusivity in the large language model space; however, this assumes that speakers of these communities desire the incorporation of their languages into such tools, which may not be the case \citep{bird-2020-decolonising}. Further, we also acknowledge that striving for zero-shot generalization to CRLs based on LLM capabilities in HRLNs undermines the need to represent CRL-specific culture and perspective of the world in LLMs \citep{hershcovich-etal-2022-challenges}. 

\section*{Acknowledgments}

We would like to thank Kaiser Sun and Vilém Zouhar for proof-reading this paper. We also thank the reviewers for their helpful feedback and suggestions.

\newpage

\bibliography{acl2023}

% VZ: let's spearate references and appendix
\clearpage

\appendix

\input{bloomz_baselines}
\input{mt0_baselines}

\section{Details of Phonological Noiser}
\label{app:phon_noiser}
\begin{figure*}[!htbp]
    \centering
    \begin{subfigure}[t]{0.48\textwidth}
        \centering
        \vspace{0pt}
        \includegraphics[width=\textwidth]{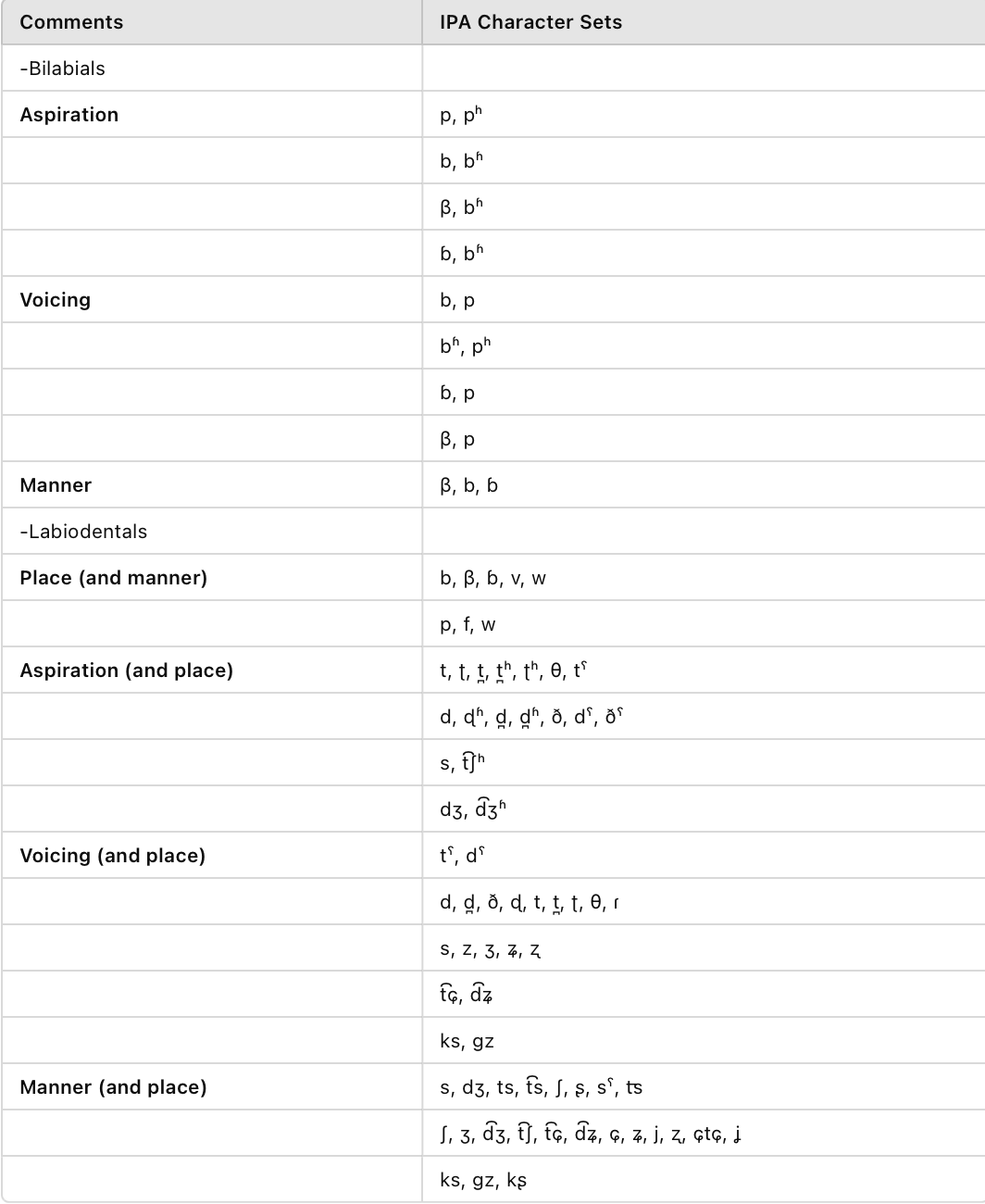}
        % \caption{Table 1}
    \end{subfigure}
    \hfill
    \begin{subfigure}[t]{0.48\textwidth}
        \centering
        \vspace{0pt}
        \includegraphics[width=\textwidth]{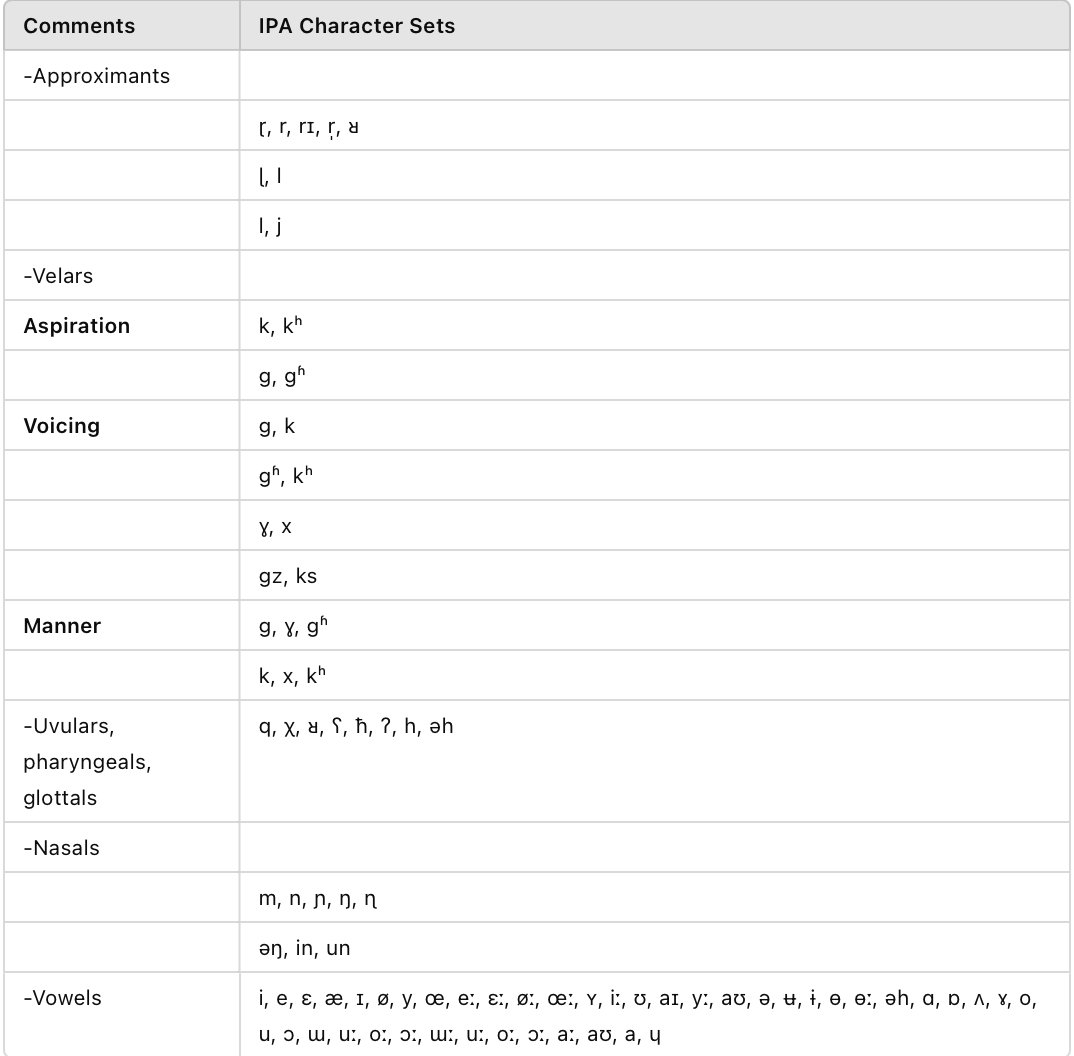}
        % \caption{Table 2}
    \end{subfigure}
    \caption{List of IPA character sets for the phonological noiser.}
    \label{fig:ipa_tables}
\end{figure*}

See \autoref{fig:ipa_tables} for the list of IPA character sets that we used in our phonological noiser described in \autoref{sec:method}. An IPA character to be noised can be transformed with uniform probability to another IPA character in any set that it belongs to.

\section{Baseline results for tasks}
\label{app:baselines}
See baseline results for \bloom{} and \mtO{} for the languages we considered in \autoref{tab:bloom_baselines} and \autoref{tab:mt0_baselines} respectively, for multilingual ARC, HellaSwag, MMLU \citep{dac2023okapi}, \flo{} \citep{nllb2022}, \xsc{} \citep{xstorycloze}, \xnli{} \citep{conneau2018xnli}, \texttt{XCopa} \citep{roemmele2011choice,ponti-etal-2020-xcopa}, \texttt{XWinoGrad} \citep{tikhonov2021heads,muennighoff2022crosslingual}, TruthfulQA \citep{lin2021truthfulqa}. We see that \bloom{} is generally better for \xsc{} and \xnli{} and work with it for the rest of our experiments. Russian and German are not included in both models but have traces in the training data as described in \citet{muennighoff2023crosslingual}; we choose to include German in our experiments as a low-resource language in \bloom{}. 

% \begin{figure}
%     \centering
%     \includegraphics[scale=0.5]{images_archive/ipa_sets1.png}
%     \caption{Caption}
%     \label{fig:enter-label}
% \end{figure}

\section{Further Experimental Details}
\label{app:experimental_details}
\input{prompts}
\subsection{Prompt Details and Variations}
\label{app:prompts}
We tried various prompts for our chosen tasks, and we note that the model performance is highly sensitive to the prompt; this has been observed in several previous studies \citep{shin2020autoprompt,gao2021making,schick2021s}. We choose a single prompting framework per task with a reasonable baseline performance in line with previous evaluations of \bloom{} \citep{muennighoff2023crosslingual}. We work in the zero-shot setting for our experiments. This is in keeping with our goal to study zero-shot generalization to unseen languages. While we note some uniform gains from including a few shots $(5-10)$ in the high-resource language, we do not study this dimension in our work.

We tried a few different prompting styles inspired by templates from Promptsource \citep{bach2022promptsource} as well as the defaults in the MLMM evaluation framework \citep{dac2023okapi} and noted considerable variation between the worst and best performing prompts (up to $15$ points for \xnli{} and $20$ points for \xsc{}). Note that for \xnli{} and \xsc{}, we see large baseline performance gains when the options are mentioned in the prompt. For \xnli{}, we also note that Prompt 3 (default) in fact requires the loglikelihood of the entire input sequence to be compared  with the corresponding labels replacing \texttt{[MASK]}, whereas the other two setups simply compare loglikehoods of the label options. See \autoref{tab:prompts}.

We also note that for \texttt{XNLI}, model performance is sensitive to the choice of word in the target language for the \texttt{entailment, neutral}, and \texttt{contradiction} labels. Interestingly, using ``No'' for the Spanish \texttt{contradiction} label results in \bloom{} loglikelihood always being highest for contradiction, possibly because it is a shared token with English, yielding near-random performance on \texttt{xnli\_es} ($33\%$)

For the translation tasks, we use Prompt $2$ for the baselines, but Prompt $1$ for the noised languages; we note that this does better than Prompt $2$ for the latter.

The above choices give rise to considerable variation in baseline performances; we work with a single setup for our experiments.

Finally, we make the choice to use English instructions for our prompts, resulting in language-mixed inputs. \bloom{} is instruction-tuned in this setup, rather than on translated prompt instructions as in the case of \texttt{mt0XXL-MT} \citep{muennighoff2023crosslingual}. We do not experiment with translated prompts to eliminate the additional complexity introduced by the quality of the translation.

\subsection{Data details}
\label{app:test_set_details}
Each evaluation is conducted over a subset of the test set consisting of $300$ samples; this is for time and compute efficiency since we conduct a large number of evaluations over combinations of task, language, noiser, and parametrization. Note that all evaluations for a given language and task are conducted over an identical subset.

All datasets used are publicly available for research use under CC BY-NC 4.0 (mARC, mHellaSwag, mMMLU), CC BY-SA 4.0 (XNLI, XStoryCloze, TruthfulQA, XCopa, FloRes200), or CC BY (XWinograd).

\subsection{Compute}
\label{app:compute}
We conduct a total of approximately $3*6*7*7 = 882$ evaluation experiments (excluding development) on NVIDIA \texttt{A100} machines, totalling about $220$ GPU hours.

\section{Results: Further details}
\label{app:details}

\begin{table*}
    \centering
    \includegraphics[scale=1.1]{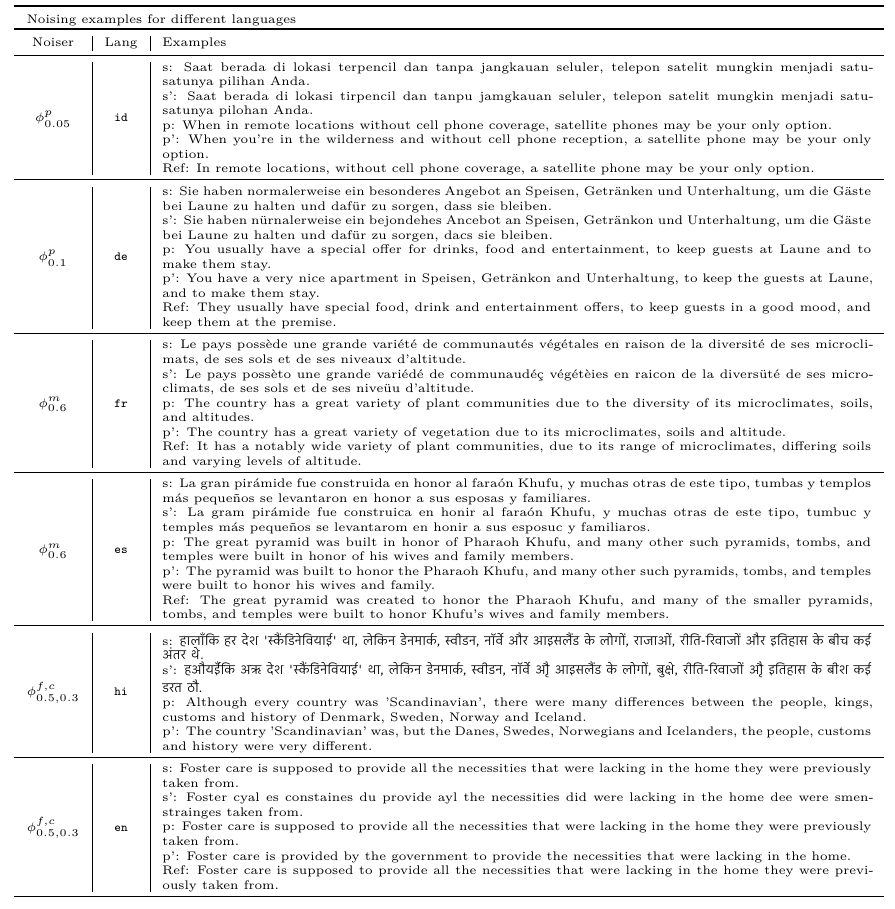}
    \caption{Examples of noising for different noisers, and model outputs for \texttt{X$\rightarrow$eng} on clean and noised source sentences. s: Source, s': Noised source, p: Prediction on source, p': Prediction on noised source, Ref: reference translation.}
    \label{tab:noiser_examples_extra}
\end{table*}

\input{all_error_mode_examples}

\subsection{Noising examples}
\label{app:noising_examples}
See \autoref{tab:noiser_examples_extra} for more examples of noiser output for certain $\theta$'s and languages. 
%We also provide the outputs for \flo{} on the clean and noised source for comparison.

\subsection{Error type examples}
\label{app:error_type_examples}

We provide an expanded version of \autoref{tab:error_types}, with an example for every mentioned error type for \texttt{es}. 
%We do not claim that is a comprehensive set of error modes; it is intended rather to be illustrative.

\input{variance}
\begin{figure}
    \centering
    \includegraphics[scale=0.7]{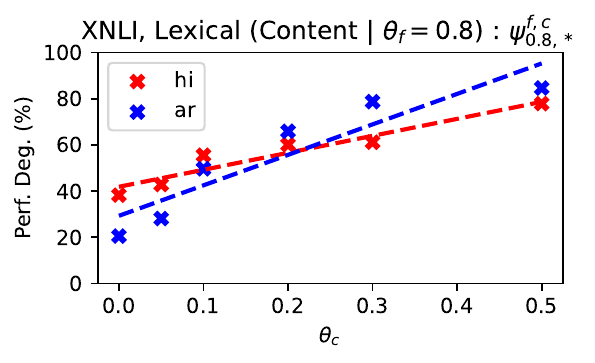}
    \caption{PD for XNLI for \texttt{hi} and \texttt{ar}, $\phi^{f,c}_{0.8,*}$, averaging over $10$ runs for each parametrization; this results in a much stabler trend for PD vs. $\theta$ as compared to using a single run as shown in \autoref{fig:results}.}
    \label{fig:meanoverruns}
\end{figure}

\subsection{Trend stability for individual languages and tasks}
\label{app:trend_stability}
In \autoref{sec:discussion}, we discuss the effect of sampling variance in PD for a given $\theta$, that appears to differ by task depending on task sensitivity to the specific words that are corrupted as opposed to the general extent of corruption in the input. We choose midrange values of $\theta_n$ for $\phi^{f,c}$, $\phi^m$, and $\phi^p$ ($\theta_f=0.5$, $\theta_c=0.3$, $\theta_m=0.5$, and $\theta_p=0.1$), and generate $10$ artificial languages with \texttt{hi} and \texttt{ar} as sources. We report standard deviation in PD for generated languages for each task in \autoref{tab:var_hi} and \autoref{tab:var_ar} for \texttt{hi} and \texttt{ar} respectively. We see that std. deviation for \flo{} is convincingly lower than for the classification tasks; this is in line with our intuition discussed in \autoref{sec:discussion}. Note that this is std. deviation in percentage PD and not actual scores: e.g., a std. deviation in PD of $10\%$ given a baseline \xnli{} score of $51$ (like for \texttt{hi}) translates to a std. deviation of $1.8$ accuracy points.\footnote{See \autoref{sec:method} for our calculation of PD.} This is low enough for our established trend to be able to provide a good ballpark estimate for the \xnli{} score for a language for which we have $\theta$.

We also recompute $\psi^{f,c}_{0.8,*}$ for \texttt{hi} and \texttt{ar} for \xnli{} (4b in \autoref{fig:results}) using means over $10$ runs per $\theta_{c}$; this combination of language, task, and noiser is motivated by the fact that the associated individual language trends appear most unstable computed over single runs per parametrization. See \autoref{fig:meanoverruns} for the trends; we observe much higher stability in the individual language trend. These findings indicate using means over several generated artificial languages in order to compute reliable trends for a single language, and using associated SD as a confidence measure in the predicted PD.

% \section{Composite noisers}

% \begin{figure}[ht]
%     \centering
%     \begin{subfigure}{0.45\textwidth}
%         \centering
%         \includegraphics[width=\linewidth]{images/composite_lex_morph_xnli.pdf}
%         \caption{Composite lexical and morphological noise for \xnli{}, for Hindi.}
%         \label{fig:composite_lex_morph_xnli}
%     \end{subfigure}
%     \hfill 
%     \begin{subfigure}{0.45\textwidth}
%         \centering
%         \includegraphics[width=\linewidth]{images/composite_lex_morph_xsc.pdf}
%         \caption{Composite lexical and morphological noise for \xsc{}, for Hindi.}
%         \label{fig:composite_lex_morph_xsc}
%     \end{subfigure}
%     \caption{}
%     \label{fig:composite_lex_morph_xnli_xsc}
% \end{figure}

% \begin{figure}[ht]
%         \centering
%         \includegraphics[width=\linewidth]{images/composite_lex_morph_xnli.pdf}
%         \caption{Composite lexical and morphological noise for \xnli{}, for Hindi.}
%     \label{fig:composite_lex_morph_xnli}
% \end{figure}

\begin{figure}[ht]
    \centering
    \begin{subfigure}{0.45\textwidth}
        \centering
        \includegraphics[width=\linewidth]{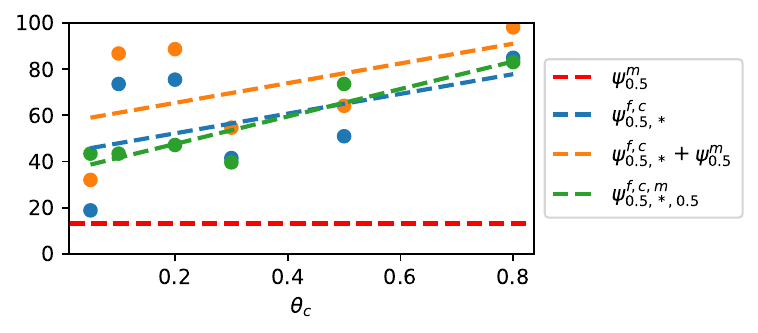}
    %     \caption{\flo{}}
    % \label{fig:composite_lex_morph_flores}
    \end{subfigure}
    \vspace{-10pt}
    \begin{subfigure}{0.45\textwidth}
        \centering
        \includegraphics[width=\linewidth]{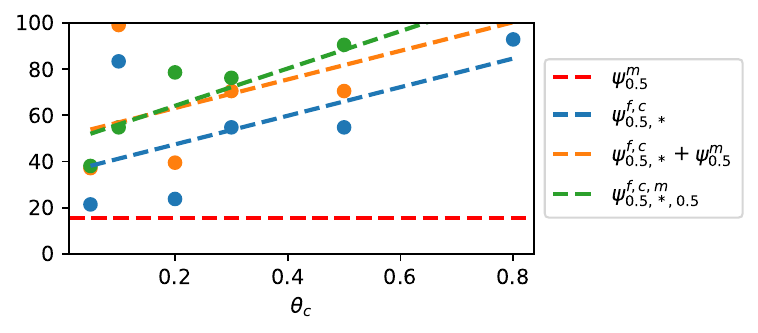}
        % \caption{\xsc{}}
        % \label{fig:composite_lex_morph_xsc}
    \end{subfigure}
    \caption{Composing $\phi^{f,c}$ and $\phi^{m}$: studying $\psi^{f,c,m}$ given  $\theta_m$ for Hindi for \xnli{} (top) and \xsc{} (bottom). $\psi^{f,c}+\psi^{m}$ shows the theoretical additive trend.}
    \label{fig:composite_lex_morph_xnli_xsc}
\end{figure}

\section{PD dynamics on composed noisers}
\label{app:pd_composed}

As discussed in \autoref{sec:discussion}, we are interested in how $\psi^{\{x,y,z\}}$ compose to give $\psi^{xyz}$ for two or more noisers, i.e. the nature of the function of PD on individual noisers that gives overall PD on composed noisers. 

\paragraph{Effect of growing lexical noise}
We describe our procedure for composing noisers and its motivation in \autoref{sec:noisers}: we apply phonological, morphological, and lexical noise in this order, independently, and allowing overwrites.

Here, we formalize why the influence of phonological and morphological noise decreases as lexical noise grows. Let's consider $\phi^{m,c}$ for simplicity; other combinations with phonological and functional noise are analogous.

Each source word $w$ independently undergoes the process of morphological and content word noising in this order, to give us the translated CRL word $w'$. At the end, we have three cases: $w'$ $\in$ $\{w, \phi_m(w), \phi_c{w}\}$. Note that $\phi_c$ overwrites $\phi_m$, so $\phi_c(\phi_m(w)) = \phi_c(w)$.

Recall that a noiser $\phi^n$ affects $w$ with probability $\theta_n$. We can then see that
\[
P(w' = w) = (1 - \theta_m)\cdot(1 - \theta_c)
\]
\[
P(w' = \phi_m(w)) = \theta_m\cdot(1 - \theta_c)
\]
\[
P(w' = \phi_c(w)) = \theta_c
\]

As $\theta_c \rightarrow 1$, $P(w' = w) \rightarrow 0$ and $P(w' = \phi_m(w)) \rightarrow 0$. This means the value of $\theta_m$ stops affecting the resulting language in the presence of high lexical noise. Note that the reverse is not true.

This effect is of course a consequence of our composition procedure allowing complete overwrites; however, this matches our intuition about linguistic variation well. Note that it is possible to nuance our noising procedure by allowing a stem overwrite while maintaining a noised suffix; we do not experiment with this idea.

\paragraph{Composite noiser curves} See \autoref{fig:composite_lex_morph_xnli_xsc} for $\psi^{f,c,m}_{0.5,*,0.5}$ for \xnli{}. We see a similar trend for \xnli{} as we saw in \autoref{fig:composite_lex_morph} for \flo{}, i.e. overall PD simply tracks the maximum individual PD (lexical in this case). However, we see that for \xsc{}, overall PD is closer to the theoretical additive trend and exceeds it for higher $\theta_c$. This difference may be indicate a dependence of the composition function of noisers on task, or the task-specific variance in PD given some $\theta^n$ (as discussed in \autoref{app:trend_stability}). We leave a more detailed investigation of noiser composition to future work.

\section{Posteriors: More details}

\subsection{Posterior computation details}
\label{app:posterior_details}
\input{posteriors}
See \autoref{tab:posteriors} for \flo{} BLEU scores on real languages, associated PD, and posteriors for all noisers computed as described in \autoref{sec:posterior_comp}. We check that using automatically aligned lexicons, which have naturally poorer quality, does not impact the posteriors too much: we verify $300$ accurate entries for the \texttt{mai-hin} and \texttt{hne-hin} silver lexicons, and obtain posteriors within $\pm 0.05$ of the posteriors computed on silver lexicons for all $\theta_n$ except for $\theta_c$ for \texttt{hne}, which is $-0.1$. $\theta_c$ is most vulnerable to being mis-estimated due to noisy alignments since it only checks for high NED. This is unlike $\theta_m$, which is computed on word pairs with the same stem, and $\theta_p$, which takes into account common phonological context on the source and target. Further, statistical word aligners are more likely to work with on very common function words, and give a roughly accurate estimate of $\theta_f$. We recommend paying attention to the quality of the lexicon for posterior computation of $\theta_c$.

\subsection{Examples of pseudo-CRLs}
\label{app:pseudo-crls}
\begin{table*}
    \centering
    \includegraphics[scale=1]{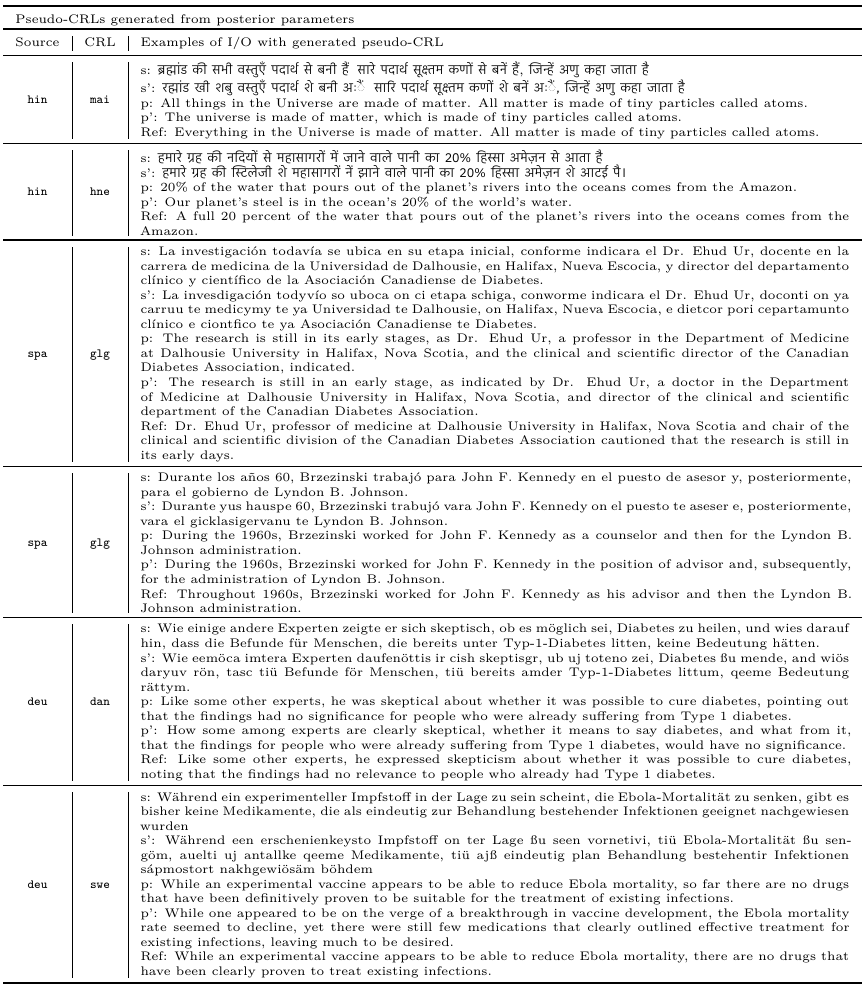}
    \caption{Examples of pseudo-CRL generated by setting noise parameters for each noiser equal to the computed posteriors for each source-CRL pair given noise type as shown in \autoref{tab:posteriors}. s: Source, s': Noised source, p: Prediction on source, p': Prediction on noised source, Ref: reference translation.}
    \label{tab:composite}
\end{table*}

Using the posteriors shown in \autoref{tab:posteriors} for a CRL relative to its HRLN, we can now generate pseudo-CRLs by composing these noise types using the procedure described in \autoref{sec:posterior_comp} (i.e. we applying $\phi^p$, $\phi^m$, $\phi^{f,c}$ in this order, independently of each other, to the HRLN). We provide examples of pseudo-CRLs generated in this manner in \autoref{tab:composite}, to illustrate noise composition in this manner.

\end{document}

%% file: error_types.tex
\begin{table*}[!ht]
\centering
    \includegraphics[scale=1]{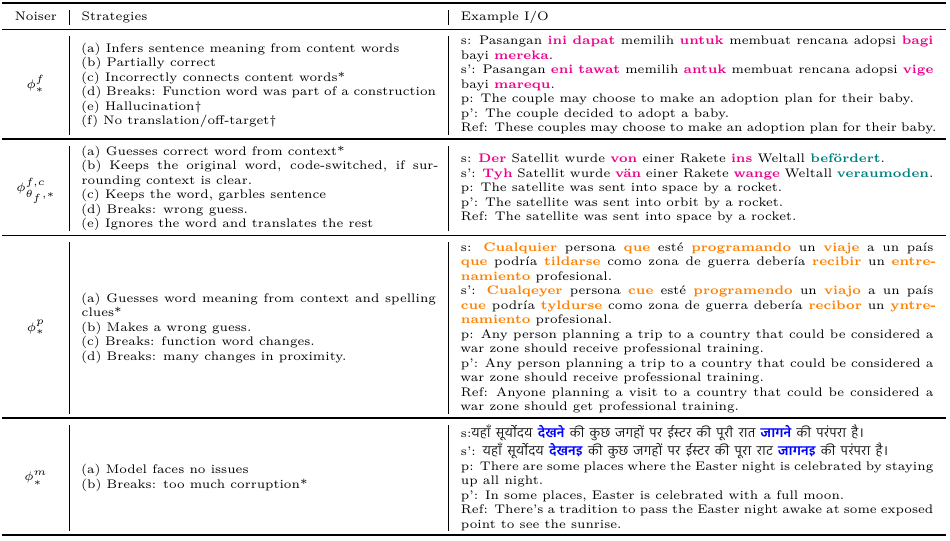}
    \caption{Output type classification for each noise type. * marks the case that the example belongs to. $\dagger$: applicable to all noisers, only listed once. Example languages from top to bottom: \texttt{id, de, es, hi}.}
    \label{tab:error_types}
\end{table*}

%% file: case_study_awa.tex
\begin{table}[ht]
    \small
    \centering
    \begin{tabular}{p{1.5cm}|c|c|p{2cm}|c}
    \toprule
     & \# & Noise type & Error class & \# \\
    \toprule
        Correct & 27  &   &   &   \\
        \midrule
        Partial & 17    & Morph & Wrong tense / person & 6 \\
        &   &   Morph+Func & Missing words & 6 \\
        &   &   Content & Missing words & 2 \\
        &   &   Morph+Func & Ungrammatical & 1 \\
        &   &   Morph & Mistranslation & 1 \\
        &   &   Phon & Missing words & 1 \\
        \midrule
        Garbled & 5 & Morph+Func & Incoherent & 5 \\
        \midrule
        Breaks & 19 & Morph+Func & Garbage & 8 \\
        &   &   Morph+Func & Keywords lost & 6 \\
        &  & Morph+Func & Repeats input & 2 \\
        &   &   Content & Garbage & 2 \\
        &   &  Content & No translation & 1 \\
        \midrule
        Hallucination & 2 & Func & Added concepts & 2 \\
        \bottomrule
    \end{tabular}
    \caption{Case study in MT error type classification for \texttt{awa-eng}. ``Noise type'' refers to the type of divergence from Hindi that causes the translation issue.}
    \label{tab:case_study}
\end{table}

%% file: bloomz_baselines.tex
\begin{table*}[!ht]
\tiny
\centering
\begin{tabular}{cccccccccc}
\toprule
 & XStoryCloze & XWinograd & XCopa & mARC & mHellaswag & mMMLU & FloRes & TruthfulQA & XNLI \\ 
\midrule
Hindi & 63.67 & - & - & 21.67 & 33.67 & 30 & 56.44 & 49.08 & 51 \\ 

Russian & 57.67 & 54.33 & - & 19.67 & 34.33 & 26 & 30.31 & 52.93 & 38.33 \\ 

Arabic & 66 & - & - & 26.33 & 32 & 32.33 & 55.32 & 48.62 & 46 \\ 

Spanish & 72.33 & - & - & 33 & 42.33 & 37.33 & 42.91 & 51.13 & 49.67 \\ 

German & - & - & - & 21 & 26 & 32 & 41.25 & 51.22 & 47.33 \\ 

Indonesian & 69.33 & - & 60.33 & 28 & 36 & 37.67 & 60 & 54.39 & - \\ 

English & 77.33 & 83.67 & - & - & - & - & 99.53 & - & 60.33 \\ 

French & - & 73.49 & - & 34.33 & 33.67 & 32.33 & 57.34 & 46.8 & 54.67 \\ 
\bottomrule
\end{tabular}
\caption{Performance of \bloom{} across different languages and tasks. }
\label{tab:bloom_baselines}
\end{table*}

%% file: mt0_baselines.tex
\begin{table*}[!ht]
\tiny
\centering
\begin{tabular}{cccccccccc}
\toprule
 & XStoryCloze & XWinograd & XCopa & mARC & mHellaswag & mMMLU & FloRes & TruthfulQA & XNLI \\ 
\midrule
Hindi & 57.3 & - & - & 28.3 & 34.6 & 30 & 52.5 & 46.4 & 39 \\ 

Russian & 57.6 & 65.3 & - & 28.6 & 36 & 32.6 & 48.1 & 46.3 & 37.1 \\ 

Arabic & 56.1 & - & - & 28.3 & 33.7 & 31.3 & 54.1 & 50.9 & 33.7 \\ 

Spanish & 59.3 & - & - & 26.6 & 37.7 & 30 & 46.1 & 45.2 & 38.6 \\ 

German & - & - & - & 25.7 & 36.3 & 22.7 & 54.1 & 44.6 & 35.6 \\ 

Indonesian & 58.3 & - & 62.5 & 28 & 39 & 30.7 & 57.5 & 43.6 & - \\ 

English & 58 & 70.3 & - & - & - & - & 99.7 & - & 50.3 \\ 
\bottomrule
\end{tabular}
\caption{Performance of \texttt{mt0XXL} across different languages and tasks. }
\label{tab:mt0_baselines}
\end{table*}

%% file: prompts.tex
\begin{table*}[ht]
    \centering
    \small
    \begin{tabular}{c|c|p{12cm}}
        \toprule
                \xnli{} & Prompt 1 & \parbox{12cm}{Suppose that the following is true:\\ \texttt{premise} \\ Can we infer that: \texttt{hypothesis}?\\ Respond with one of the following words: \texttt{ENTAILMENT\_LABEL}, \texttt{CONTRADICTION\_LABEL}, \texttt{NEUTRAL\_LABEL}.} \\
                
    \cmidrule(lr){2-3}
    
        & Prompt 2 & \parbox{12cm}{Suppose that the following is true:\\ \texttt{premise} \\ Can we infer that: \texttt{hypothesis}? Yes, no, or maybe? \\ Respond in the target language.} \\
        \cmidrule(lr){2-3}
        & Prompt 3$^{*}$ & \parbox{12cm}{\texttt{premise, QUESTION\_WORD}? [MASK], \texttt{hypothesis}} \\
        % \cmidrule(lr){2-3}
        \midrule

        \texttt{XStoryCloze} & Prompt 1 & \parbox{12cm}{What is a possible continuation for the following story ? \\ \\ \texttt{sentence\_1} \\
        \texttt{sentence\_2} \\
        \texttt{sentence\_3} \\
        \texttt{sentence\_4} \\ \\
        Choose from the following options: \\ 
        \texttt{option\_1} \\ 
        \texttt{option\_2}} \\
        \cmidrule(lr){2-3}
        & Prompt 2 & \parbox{12cm}{\texttt{sentence\_1} \texttt{sentence\_2} \texttt{sentence\_3} \texttt{sentence\_4} \\
    What is a possible continuation for the story given the following options ? \\ \\
     - \texttt{option\_1} \\
     - \texttt{option\_2}} \\  
     \cmidrule(lr){2-3}
      &  Prompt 3 & \parbox{12cm}{Choose the best continuation of this story: 
       \texttt{sentence\_1} \\
        \texttt{sentence\_2} \\
        \texttt{sentence\_3} \\
        \texttt{sentence\_4}} \\
                
    \midrule

    \flo{} & Prompt 1 & \parbox{12cm}{Translate from a dialect of \texttt{<HRLN>} into English} \\
    \cmidrule(lr){2-3}
    & Prompt 2 & \parbox{12cm}{Translate from \texttt{<HRLN>} into English} \\
    \cmidrule(lr){2-3}
    & Prompt 3 & \parbox{12cm}{Translate into English :} \\

    \bottomrule

    \end{tabular}
    \caption{Our attempted prompts. $^{*}$\texttt{[MASK]} is filled with each of the three possible labels, and the model choice is computed using loglikelihood over the entire sequence.}
    \label{tab:prompts}
\end{table*}

%% file: all_error_mode_examples.tex
\begin{table*}[!ht]
    \centering
    \tiny
    \begin{tabular}{c|p{4cm}|p{10cm}}
    \toprule
    \multicolumn{3}{l}{Examples for all error modes} \\
    \toprule
    Noiser & Strategies & Example I/O \\
    \midrule
    $\phi^{f}_{*}$ & \parbox{4cm}{(a) Infers sentence meaning from content words} & \parbox{10cm}{s: Al parecer, las cabras fueron domesticadas, por primera vez, hace unos 10 000 años, en los montes Zagros, en Irán.\\
    s': Al parecer, luc cabras fiaom domesticadas, por primera vez, hace enes 10 000 años, an los montes Zagros, an Irán. \\
    p: Apparently, goats were first domesticated about 10,000 years ago in the Zagros Mountains in Iran. \\
    p': It seems that the first domesticated goats were bred in the Zagros Mountains of Iran about 10,000 years ago. \\
    Ref: Goats seem to have been first domesticated roughly 10,000 years ago in the Zagros Mountains of Iran.}\\
    \midrule
    
    & \parbox{4cm}{(b) Partially correct} & \parbox{10cm}{s: Los esfuerzos para hallar el lugar del accidente deben lidiar con el mal tiempo y el terreno escarpado. \\
    s': Los esfuerzos pea hallar al lugar del accidente cebyn lidiar kom al ah tiempo i al terreno escarpado. \\
    p: The efforts to find the crash site must contend with bad weather and rugged terrain. \\
    p': The efforts were made to find the place of the accident, but the terrain was too rough. \\
    Ref: Efforts to search for the crash site are being met by bad weather and harsh terrain.} \\

    \midrule
    
    & \parbox{4cm}{(c) Incorrectly connects content words*} & \parbox{10cm}{s: Las manifestaciones, en ocasiones violentas, fueron provocadas por el hecho de que no se llevan adelante elecciones, en algunos casos desde el año 2011. \\
    s': Luc manifestaciones, an ocasiones violentas, fiaom provocadas por al hecho de guu no ze llevan adelante elecciones, an olgones casos ceztu al año 2011. \\
    p: The protests, sometimes violent, were sparked by the fact that elections are not held in some cases since 2011. \\
    p': In 2011, there were violent protests, sometimes triggered by the failure to hold elections. \\
    Ref: The sometimes-violent protests were triggered by failure to hold elections, some due since 2011.} \\

    \midrule
    
    & \parbox{4cm}{(d) Breaks: Function word was part of a construction} & \parbox{10cm}{s: Sin perjuicio de cuán mansos puedan lucir, lo cierto es que los bisones, los uapatíes, los alces, los osos y prácticamente todos los animales grandes pueden se agresivos.\\
    s': Sin perjuicio de ceám mansos piedan lucir, li cierto os guu los bisones, los uapatíes, los alces, los osos i prácticamente dodus los animales grandes pieden ze agresivos.\\
    p: No matter how docile they may look, bears, bison, moose, elk, bears, and nearly all large animals can be aggressive.\\
    p': Without prejudice to the fact that bison, moose, elk, bears, and nearly all large animals can be aggressive, it is true that the bisons, moose, elk, bears, and nearly all large animals can be very docile.\\
    Ref: No matter how docile they may look, bison, elk, moose, bears, and nearly all large animals can attack.}\\

    \midrule

    & \parbox{4cm}{(e) Off-target} & \parbox{10cm}{s: Se han rescatado varios rehenes y, hasta ahora, se ha confirmado que al menos seis han muerto.\\
    s': Ze han rescatado parius rehenes i, hosta ahora, ze he confirmado guu al menos seis han muerto.\\
    p: Several hostages have been rescued, and it is confirmed that at least six have died so far.\\
    p': Spanish phrase: Ze han rescatado parius rehenes i, hosta ahora, ze he confirmado guu al menos seis han muerto.\\
    Ref: Several hostages have been rescued and least six have been confirmed dead so far.}\\
    
    \midrule
    
    $\phi^{f,c}_{\theta_f,*}$ & \parbox{4cm}{(a) Guesses correct word from context} & \parbox{10cm}{s: Todo en el Universo está hecho de materia, \textbf{compuesta} por \textbf{partículas} \textbf{pequeñas} denominadas átomos.\\
    s': Todo en el Universo está hecho de materia, \textbf{tespolaci} por \textbf{piamplesc} \textbf{obleyón} denominadas átomos.\\
    p: Everything in the Universe is made of matter, composed of tiny particles called atoms.\\
    p': Everything in the Universe is made of matter, which is made of tiny particles called atoms.\\
    Ref: Everything in the Universe is made of matter. All matter is made of tiny particles called atoms.} \\

    \midrule

    & \parbox{4cm}{(b) Keeps the original word, code-switched, if sur- rounding context is clear} & \parbox{10cm}{s: Los rasgos que distinguen a una subcultura pueden ser lingüísticos, estéticos, sexuales, geográficos o estar relacionados con la religión o la política, o una mezcla de factores.\\
    s': Los rasgos que distinguen a una calincio pueden ser teleamplinempal, estéticos, sexuales, esolaridalla o estar relacionados con la religión o la política, o una mezcla de factores.\\
    p: The characteristics that distinguish a subculture can be linguistic, aesthetic, sexual, geographical, religious, or political, or a combination of factors.\\
    p': The characteristics that distinguish a calincio can be teleamplinempal, aesthetic, sexual, esolaridalla, or related to religion or politics, or a mixture of factors.\\
    Ref: The qualities that determine a subculture as distinct may be linguistic, aesthetic, religious, political, sexual, geographical, or a combination of factors.} \\

    \midrule
    
    & \parbox{4cm}{(c) Keeps the word, garbles sentence} & \parbox{10cm}{s: El satélite en el espacio recibe la llamada y, luego, la refleja de vuelta casi de forma instantánea.\\
    s': El devasalv en el espacio recibe la llamada y, vircap, la refleja de vuelta apases de bharítu instantánea.\\
    p: The satellite in space receives the call and then reflects it back almost instantly.\\
    p': The devasalv in space receives the call and, vircap, reflects it back to the instantaneous bharítu.\\
    Ref: The satellite in space gets the call and then reflects it back down, almost instantly.} \\

    \midrule
    & \parbox{4cm}{(d) Breaks: wrong guess} & \parbox{10cm}{ s: Los entomólogos emplean el término insecto parásito en un sentido formal para referirse a este grupo de artrópodos.\\
    s': Los entomólogos ceradida el cataciónit insecto ingaren en un sintaut formal para referirse a este scomp de artrópodos.\\
    p: The entomologists use the term insect parasite in a formal sense to refer to this group of arthropods.\\
    p': The entomologists use the term insectivore to refer to this group of arthropods.\\
    Ref: The term bug is used by entomologists in a formal sense for this group of insects.}\\

    \midrule
    
    & \parbox{4cm}{(e) Ignores the word and translates the rest} & \parbox{10cm}{s: Hershey y Chase insertaron su propio ADN en una bacteria usando fagos, o virus.\\
    s': Hershey y Chase insertaron su propio Adn en una resabajectoma usando capandil, o virus.\\
    p: Hershey and Chase inserted their own DNA into a bacterium using phages, or viruses.\\
    p': Hershey and Chase inserted their own Adn into a somatic cell using capandil, or virus.\\
    Ref: Hershey and Chase used phages, or viruses, to implant their own DNA into a bacterium.} \\
    \midrule
    \bottomrule
    \end{tabular}
    \caption{Examples of each error mode for \texttt{es}. Continued below.}
    \label{tab:all_error_modes1}
\end{table*}

\begin{table*}[!ht]
    \centering
    \tiny
    \begin{tabular}{c|p{4cm}|p{10cm}}
    \toprule
    \multicolumn{3}{l}{Examples for all error modes} \\
    \toprule
    Noiser & Strategies & Example I/O \\
    \midrule
    
    $\phi^{p}_{*}$ & \parbox{4cm}{(a) Guesses word meaning from context and spelling clues} & \parbox{10cm}{s: El informe es sumamente crítico con prácticamente cada aspecto de la política vigente del poder ejecutivo en Irak, y apela a un cambio inmediato de dirección.\\
    s': Ey informe es sumamenty crítico con prácticamente cada aspecto de la política vigenty del pider eyetutivo ym Irak, e apela a un camvuo inmediato de dirección.\\
    p: The report is highly critical of almost every aspect of the present executive policy in Iraq, and urges an immediate change of direction.\\
    p': The report is highly critical of almost every aspect of the present policy of the U.S. towards Iraq, and it calls for an immediate change of direction.\\
    Ref: The Report is highly critical of almost every aspect of the present policy of the Executive towards Iraq and it urges an immediate change of direction.} \\

    & \parbox{4cm}{(b) Makes a wrong guess.} & \parbox{10cm}{s: La investigación en el campo de la IA supone el desarrollo de máquinas a fin de automatizar tareas que requieren un comportamiento inteligente.\\
    s': La investigación ym ul campo de la IA sopone ul desarrolyo de máquinas a fin de audymatizor caeas cue reqeyerem un comportamiento inteligente.\\
    p: Research in the field of AI involves the development of machines to automate tasks that require intelligent behavior.\\
    p': The research in the field of AI involves the development of machines to automate tasks so that machines can exhibit intelligent behavior.\\
    Ref: Research in AI involves making machines to automate tasks that require intelligent behavior.}\\

    \midrule

    & \parbox{4cm}{(d) Breaks: many changes in proximity.} & \parbox{10cm}{s: No olvide que, básicamente, usted está visitando un lugar que ofició de fosa común y que también es un sitio de un significado prácticamente invaluable para una parte importante de la población del mundo.\\
    s': No ylvide que, básicamente, ustat está visitando un lugar cue ofició de fosa común e cue también es un sitio de un signifijado prácticamente imvaluable para una party importanty de la población del mundo.\\
    p: Don't forget that, basically, you're visiting a place that served as a mass grave and that it is also a place of essentially invaluable significance to a significant part of the world's population.\\
    p': No ylvide that, basically, ustat is visiting a place that was a fosa común and also a place that has a practically invaluable meaning for a party importanty of the population of the world.\\
    Ref: Please remember that you are essentially visiting a mass grave site, as well as a site that has an almost incalculable meaning to a significant portion of the world's population.}\\

    \midrule
    & \parbox{4cm}{(e) Hallucination} & \parbox{10cm}{s: Es tradición pasar la noche de Pascua en vela en algún sitio expuesto para contemplar la salida del sol.\\
    s': Es tradición fasa la noche de Paszua an vyla an algún sutio uxpaesdo fary comtemfla la caleda del sol.\\
    p: It is tradition to spend the night of Easter awake at some exposed place to watch the sunrise.\\
    p': It is tradition to make the night of Pascuas by lighting a bonfire in the yard.\\
    Ref: There's a tradition to pass the Easter night awake at some exposed point to see the sunrise.}\\

\midrule

    $\phi^{m}_{*}$ & \parbox{4cm}{(a) Model faces no issues} & \parbox{10cm}{s: Montevideo se ubica en los subtrópicos, con frecuentes temperaturas superiores a +30° C durante el verano.\\
    s': Montevidyo se ubiga en los subtrópicos, con frecuentec temperaturaz superiorec a +30° C durante el verani.\\
    p: Montevideo is located in the subtropics, with frequent temperatures above +30°C during the summer.\\
    p': Montevideo is in the subtropics, with frequent temperatures above +30°C during the summer.\\
    Ref: Montevideo is in the subtropics; in the summer months, temperatures above +30°C are common.}\\
    \midrule
    & \parbox{4cm}{(b) Breaks: too much corruption*}& \parbox{10cm}{s: Il est de tradition de passer la nuit de Pâques éveillé à un endroit à découvert pour voir le lever du soleil.\\
    s': Il est de traditiin de pasjer la nuèt de Pâques éveillé à un endroèt à découvert pour vâyr le levir du soleel.\\
    p: It is traditional to stay up all night on Easter Sunday to see the sunrise.\\
    p': Traditionally, it is custom to wake up at dawn on Easter Sunday to see the sunrise at a place of worship.\\
    Ref: There's a tradition to pass the Easter night awake at some exposed point to see the sunrise.}\\
    \bottomrule
    \end{tabular}
    \caption{Continued from \autoref{tab:all_error_modes1}: Examples of each error mode for \texttt{es}.}
    \label{tab:all_error_modes2}
\end{table*}

%% file: variance.tex
\begin{table}[ht]
\centering
\small
\begin{tabular}{lcccc}
\toprule
            & $\phi^{f,c}_{0.5,0.3}$ & $\phi^m_{0.5}$ & $\phi^p_{0.1}$ & \textbf{Task Avg.} \\
\midrule
X->eng      & 4.4  & 2.6  & 4.6  & 3.9  \\
XNLI        & 18.1 & 9.7  & 17.0 & 14.9 \\
XStoryCloze & 16.5 & 10.7 & 11.2 & 12.8 \\
\textbf{Noiser Avg.}  & 13.0 & 7.7  & 10.9 & -  \\
\bottomrule
\end{tabular}
\caption{Std. dev. of PD\% over $10$ artificial languages generated by a given noiser for each task, for \texttt{hi}}
\label{tab:var_hi}
\end{table}

\begin{table}[ht]
\centering
\small
\begin{tabular}{lcccc}
\toprule
            & $\phi^{f,c}_{0.5,0.3}$ & $\phi^m_{0.5}$ & $\phi^p_{0.1}$ & \textbf{Task Avg.} \\
\midrule
X->eng      & 2.8  & 2.0  & 6.9  & 3.9  \\
XNLI        & 9.3  & 10.9 & 6.5  & 8.9  \\
XStoryCloze & 14.3 & 14.6 & 20.3 & 16.4 \\
\textbf{Noiser Avg.}  & 8.8  & 9.2  & 11.2 & -  \\
\bottomrule
\end{tabular}
\caption{Std. dev. of PD\% over $10$ artificial languages generated by a given noiser for each task, for \texttt{ar}}
\label{tab:var_ar}
\end{table}

% lexical: 0.5, 0.3 func, content
% morph: 0.5
% phon: 0.1

%% file: posteriors.tex
\begin{table}[ht]
\centering
\tiny
\begin{tabular}{ll|cccc|cc}
\toprule
Source & CRL & $\theta^{c}$ & $\theta^{f}$ & $\theta^m$ & $\theta^p$ & BLEU & PD (\%) \\
\midrule
\texttt{hin} & \texttt{hin} & 0 & 0 & 0 & 0 & 56.44 & 0 \\
& \texttt{awa} & 0.15 & 0.67 & 0.26 & 0.05 & 37.03 & 34.39 \\
& \texttt{bho} & 0.24 & 0.79 & 0.32 & 0.07 & 32.38 & 42.63 \\
& \texttt{hne} & 0.18 & 0.67 & 0.24 & 0.05 & 33.24 & 41.11 \\
& \texttt{mag} & 0.14 & 0.7 & 0.26 & 0.05 & 41.47 & 26.52 \\
& \texttt{mai} & 0.2 & 0.81 & 0.34 & 0.04 & 28.4 & 49.68 \\
\midrule
\texttt{ind} & \texttt{ind} & 0 & 0 & 0 & 0 & 60 & 0 \\
& \texttt{zsm} & 0.19 & 0.46 & 0.13 & 0.06 & 53.01 & 11.65 \\
\midrule
\texttt{spa} & \texttt{spa} & 0 & 0 & 0 & 0 & 42.91 & 0 \\
& \texttt{glg} & 0.22 & 0.71 & 0.2 & 0.11 & 47.01 & -9.55 \\
\midrule
\texttt{fra} & \texttt{fra} & 0 & 0 & 0 & 0 & 57.34 & 0 \\
& \texttt{oci} & 0.57 & 0.88 & 0.73 & 0.09 & 38.4 & 33.03 \\
\midrule
\texttt{deu} & \texttt{deu} & 0 & 0 & 0 & 0 & 41.25 & 0 \\
& \texttt{dan} & 0.5 & 0.98 & 0.71 & 0.1 & 16.37 & 60.32 \\
& \texttt{isl} & 0.75 & 0.99 & 0.68 & 0.15 & 4.11 & 90.04 \\
& \texttt{swe} & 0.56 & 0.99 & 0.7 & 0.1 & 16.7 & 59.52 \\
\midrule
\texttt{arb} & \texttt{arb} & 0 & 0 & 0 & 0 & 55.32 & 0 \\ 
& \texttt{acm} & 0.09 & 0.32 & 0.08 & 0.03 & 24.17 & 56.31 \\ 
& \texttt{acq} & 0.06 & 0.25 & 0.04 & 0.04 & 46.76 & 15.47 \\ 
& \texttt{aeb} & 0.2 & 0.43 & 0.11 & 0.05 & 43.55 & 21.28 \\ 
& \texttt{ajp} & 0.21 & 0.55 & 0.15 & 0.04 & 38.25 & 30.86 \\ 
& \texttt{apc} & 0.21 & 0.64 & 0.18 & 0.04 & 44.41 & 19.72 \\ 
& \texttt{ars} & 0.02 & 0.02 & 0.01 & 0.05 & 48.36 & 12.58 \\ 
& \texttt{ary} & 0.32 & 0.6 & 0.12 & 0.03 & 50.16 & 9.33 \\ 
& \texttt{arz} & 0.19 & 0.5 & 0.1 & 0.04 & 33.05 & 40.26 \\ 
\bottomrule
\end{tabular}
\caption{Posteriors for related languages, BLEU scores for \flo{}, and corresponding PD.}
\label{tab:posteriors}
\end{table}